\documentclass[11pt]{article}

\PassOptionsToPackage{numbers, compress}{natbib}

\usepackage[final,eandd]{neurips_2026}

\usepackage{amsmath}
\usepackage{amssymb}
\usepackage{graphicx}
\usepackage{booktabs}
\usepackage{algorithm}
\usepackage{algorithmic}
\usepackage{subcaption}
\usepackage{multirow}
\usepackage{longtable}
\usepackage{xcolor}
\usepackage{hyperref}
\usepackage{url}
\usepackage{placeins}
\usepackage{wrapfig}
\usepackage{tikz}
\usetikzlibrary{positioning, fit, arrows.meta, decorations.pathreplacing}

\usepackage{todonotes}

\usepackage{listings}
\lstdefinelanguage{Rust}{
  keywords={fn, let, mut, pub, struct, impl, enum, match, self, use, mod, trait, where, for, in, if, else, return, type, const, static, ref, move, Box, Vec, Option, Some, None},
  keywordstyle=\color{blue}\bfseries,
  ndkeywords={f64, usize, bool, i32, u64, String},
  ndkeywordstyle=\color{teal}\bfseries,
  sensitive=true,
  comment=[l]{//},
  morecomment=[s]{/*}{*/},
  commentstyle=\color{gray}\itshape,
  stringstyle=\color{red},
  morestring=[b]",
  basicstyle=\ttfamily\small,
  breaklines=true,
  showstringspaces=false,
  tabsize=4
}
\newcommand{\rustknn}{\textsc{RustKNN}}
\newcommand{\covdist}{\textsf{covdist}}
\newcommand{\sepdist}{\textsf{sepdist}}
\newcommand{\maxdist}{\textsf{maxdist}}

\title{KNN Implementation Details Can Dramatically Change Performance: An Example from Cover Trees}

\author{%
  Amol Khanna \\
  CrowdStrike \\
  \texttt{amol.khanna@crowdstrike.com} \\
  \And
  Edward Raff \\
  CrowdStrike \\
  University of Maryland, Baltimore County \\
  Syracuse University \\
  \texttt{edward.raff@crowdstrike.com} \\
  \texttt{raff.edward@umbc.edu} \\ 
  \texttt{edraff@syr.edu}\\
}

\begin{document}

\maketitle

\begin{abstract}
% Cover trees provide exact nearest neighbor search in arbitrary metric spaces,
% yet practical performance depends on interactions between query strategy,
% neighbor count~$k$, base parameter, and micro-optimizations that are poorly
% understood because existing implementations span different languages and
% codebases.
% We present a systematic empirical study of these interactions using
% \rustknn{}, a unified Rust implementation ($\sim$13{,}000 lines)
% covering all four algorithms from~\citet{izbicki2015faster} plus dual-tree
% traversal.
Nearest neighbor search algorithms are widely used and have a long history, but we find that key implementation details are seldom reviewed in articles about them. These details can materially impact performance, and thus change the conclusions one makes about a given algorithm's superiority. In this article, we highlight these impacts by reviewing the Cover Tree as an example, producing \rustknn{}, a unified Rust implementation of three construction variants of the Cover Tree with three query modes. 
Our analysis reveals:
(1)~the optimal query strategy depends on $k$ and on the dataset---single-tree search is fastest at $k{=}1$ on three of our four largest datasets, while dual-tree search is fastest at $k{\ge}5$ on the three largest---and choosing the wrong query mode can cost up to a $7.5\times$ factor in runtime;
(2)~the cover tree base parameter, set to 2 in theory and 1.3 in practice, is not a safe default for dual-tree search: 1.3 is optimal on only one of seven datasets, and on covtype it is $1.6\times$ slower than the optimum;
(3)~among five micro-optimizations, only early-exit distance computation has a consistent effect, and that effect grows with dimension
(none at $d{=}32$, $1.31\times$ at $d{=}54$, and $1.49\times$ at $d{=}784$);
and (4)~all-nearest-neighbor and held-out benchmarks yield different
implementation rankings.
All main timing tables report medians over 20 independent allocations with bootstrap 95\% confidence intervals.
\rustknn{} is open source at \url{https://github.com/CrowdStrike/RustKNN}.
\end{abstract}

%% ===========================================================================
\section{Introduction}
\label{sec:introduction}

Nearest neighbor algorithms, and data structures to accelerate/improve them, exist in many forms. There are algorithms for pruning the search space like the KD-tree~\cite{moore-tutorial}, sketches for calculating approximations of distance functions faster~\citep{Ioffe:2010:ICS:1933307.1934593}, and more. This line of research has been highly impactful in real-world systems, from information retrieval to clustering/visualization and even in Large Language Models via retrieval-augmented generation and vector databases. While \emph{approximate} methods have dramatically gained popularity in recent years, exact metric-space $k$-NN remains necessary when the distance itself is the object of interest or approximate recall is unacceptable---for example, in cases like anomaly detection over raw features, protein-structure comparison under graph kernels, or retrieval under Wasserstein distance. However, when we survey papers (see Appendix~\ref{sec:app_litsurvey}) in this broad area, we find that few consider basic details, such as distinguishing between a hold-out set's query performance and the original training data (26\%). Fewer still report \emph{both} distance-comparison counts and runtime (15\%), even though the two are not always correlated, document micro-optimization details (11\%)---the low-level implementation choices we isolate in Section~\ref{sec:ablation}---or batch vs.\ single-instance queries (10\%).

We argue that papers evaluating the performance of nearest neighbor dependent systems need to document and evaluate these factors carefully, or give a justified reason why only a subset matters for their use case. We support that argument by developing \rustknn{} as an example, re-evaluating the methods behind the seminal cover tree data structure---a structure with multiple proposed variants and one of the few with both empirical success and theoretical grounding for reducing nearest neighbor queries. We re-implement a cache-efficient cover tree with three query settings (single, batch, and dual-tree). Working in a compiled language allows both effective runtime evaluation and easy isolation of six micro-optimizations used in the literature (five of which we ablate), along with poorly documented nuances of the cover tree ``base'' parameter, which to our knowledge has never been independently tested.

% Despite algorithmic improvements, a gap remains in \emph{evaluating}
% cover tree performance: existing implementations span different languages
% (C++ and Haskell), implement different tree construction and query algorithms, use different hyperparameters, and use
% different micro-optimizations, making it impossible to conduct controlled
% comparisons. A controlled comparison is essential to benchmark the optimal performance of cover trees. To this end, we present \rustknn{}, a unified Rust implementation of cover trees
% ($\sim$13{,}000 lines) that serves as a controlled experimental
% platform.
% The library implements five tree construction variants (simplified, nearest ancestor,
% packed, nearest ancestor packed, parallel-merged), three query algorithms (single-tree, batch
% single-tree, dual-tree), and six compile-time optimization
% toggles---all sharing the same distance trait, node structure, and
% $k$-NN state machine.
% This design enables the first systematic study of how cover tree
% performance depends on the interaction of these factors.
Our contributions are:
\begin{enumerate}
  \item \textbf{Optimization ablation.}
    We quantify the individual and cumulative impact of five
    micro-optimizations via feature-flag ablation, revealing that
    early-exit distance computation is the only one with a consistent effect,
    and that its benefit grows with dimension (none on corel, $d{=}32$; $1.31\times$ on covtype, $d{=}54$; and
    $1.49\times$ on mnist, $d{=}784$), while the
    traversal-level optimizations sit within about 10\% of $1.0\times$
    (Section~\ref{sec:ablation}).
  \item \textbf{Query mode $\times$ $k$ interaction.}
    We show that the optimal query strategy depends on both~$k$ and the
    dataset: single-tree search is fastest at $k{=}1$ on mnist yet $3.3\times$
    slower than dual-tree search at $k{=}10$ on that same dataset, and
    choosing the wrong query mode costs up to $7.5\times$ in runtime
    (Section~\ref{sec:mode_k}).
  \item \textbf{Base parameter analysis.}
    We show that the conventional base~1.3 is not a safe default for
    dual-tree search: the optimum is~2.0 on four of seven datasets at
    $k{=}1$ (five at $k{=}10$) and~3.0 on the two smallest, and the default is
    $1.6\times$ slower than the optimum on covtype, the largest dataset
    (Section~\ref{sec:base_analysis}).
  \item \textbf{All-NN vs.\ held-out divergence.}
    We demonstrate that the standard all-nearest-neighbor benchmark
    produces different implementation rankings than held-out evaluation,
    with implications for how the community evaluates nearest neighbor
    methods (Section~\ref{sec:heldout_divergence}).
\end{enumerate}

% Table~\ref{tab:allnn} previews the performance landscape.
% At $k{=}1$, no single query mode dominates: batch single-tree wins on 7 of
% 12~datasets, dual-tree wins on corel, and single-tree wins on mnist.
% At $k{=}10$, batch single-tree dominates decisively, achieving up to
% $111\times$ over the~\citet{beygelzimer2006cover} reference C++ implementation on mnist---a crossover
% that Table~\ref{tab:k_scaling} traces across five values of~$k$.
% These interactions, invisible when comparing implementations across
% different codebases, are the focus of our analysis.
In doing so, we find that no single method dominates; the results are dependent on dataset and query settings, and thus far more complicated and involved than the literature has previously implied. This complexity exists within just the cover tree literature alone, and we find it instructive to the questions we may have of the broader literature. 

Sections~\ref{sec:related}--\ref{sec:experiments} cover related work, background, our implementation, and the evaluation.

%% ===========================================================================
\section{Related Work}
\label{sec:related}
%% ===========================================================================

\textbf{Nearest neighbor data structures.}
Spatial indexing methods for nearest neighbor search can be organized along two
axes: exact versus approximate, and coordinate-based versus general metric
space.
Among exact methods for coordinate spaces,
KD-trees~\citep{friedman1977algorithm} achieve $O(\log n)$ query time in low
dimensions but degrade to $O(n)$ as dimensionality grows, and their
axis-aligned splits tie them to a coordinate representation;
ball trees~\citep{omohundro1989five} improve on KD-trees through hierarchical
ball decomposition, extending efficient search to moderate dimensions; because
they partition by distance to centers rather than along axes, they apply to any
proper metric.
For general metric spaces, vantage-point
trees~\citep{yianilos1993data} use binary partitions around single pivot points
but offer only $O(n)$ worst-case query time.
Navigating nets~\citep{krauthgamer2004navigating} introduced the theoretical
framework of the doubling dimension $c$ (also called the ``expansion constant''), which
cover trees~\citep{beygelzimer2006cover} made practical with an $O(c^6 \log n)$
query algorithm, later improved by~\citet{izbicki2015faster} with simplified and
nearest ancestor variants(though a history of nuanced theoretical runtime errors and fixes~\cite{Raff2018_metric_index,elkin_new_2023,elkin_counterexamples_2022,Curtin2015} is beyond this study), tree merging for parallel construction, and
cache-optimized packing.
On the approximate side, locality-sensitive hashing
(LSH)~\citep{indyk1998approximate} provides sublinear search but requires
metric-specific hash families, while hierarchical navigable small world graphs
(HNSW)~\citep{malkov2020efficient} achieve state-of-the-art recall--speed
tradeoffs for approximate search but offer no exactness guarantees.
Cover trees are a \emph{practical} structure combining exact search, arbitrary
metric spaces, and query complexity parametrized by intrinsic rather than
ambient dimensionality. They are not alone in that combination---navigating
nets, cited above, share it---but they are the variant with the broadest
empirical track record, which is why we take them as our case study.

\textbf{Implementations and libraries.}
The original cover tree C++ reference implementation~\citep{beygelzimer2006cover},
last updated in 2014, is single-threaded with distance-level optimizations
(batched computation, aligned memory) but no tree-level cache packing.
mlpack~\citep{curtin2013mlpack} provides the most actively maintained cover tree
implementation, including dual-tree traversal and extensive test coverage.
FLANN~\citep{muja2014scalable} offers randomized KD-trees and is widely used in
robotics and computer vision, though it targets coordinate spaces only.
Among approximate and GPU-accelerated systems,
FAISS~\citep{johnson2019billion} scales to billions of vectors with
quantization-based search on GPUs;
Annoy~\citep{bernhardsson2013annoy} uses random projection trees with
memory-mapped files for production use;
and ScaNN~\citep{guo2020accelerating} employs learned anisotropic vector
quantization.
Existing cover tree implementations span C++, Haskell, and Java~\cite{JMLR:v18:16-131,Hall2009} (e.g., JSAT \& WEKA) -but no two of them implement the
same set of variants, query modes, and micro-optimizations, and none is
Rust-native.
Critically, cross-language and cross-codebase comparisons confound
algorithmic effects (e.g., nearest ancestor vs.\ simplified) with
implementation effects (e.g., compiler optimizations, memory layout,
distance computation strategy), making it difficult to attribute
performance differences to their true cause. Our work focuses on cover trees
specifically, but we believe the insight of delineating implementation details
applies to other search structures and can reveal new research directions;
Section~\ref{sec:beyond_ct} gives direct evidence for a second structure.
\rustknn{} is the first Rust-native cover tree, integrating directly with the
Rust data-processing ecosystem without foreign-function interface overhead;
more importantly for this study, it implements all variants and query modes
in a single codebase, enabling controlled comparisons that isolate one
factor at a time. These relate to issues in both maintainability and replicability~\cite{raff_what_2025}, to which a larger literature is relevant but beyond our scope~\cite{arvan_reproducibility_2022,10.1145/3324884.3416545,Kanewala2014,Raff2022a,Carver2007,gundersen_machine_2022,Raff2020c,Raff2022a}.

\textbf{Rust in the ML ecosystem.}
Rust adoption in machine learning infrastructure has accelerated for
performance-critical components.
Hugging Face tokenizers~\citep{tokenizers} use a Rust backend;
candle~\citep{candle} and burn~\citep{burn} provide Rust deep learning
frameworks;
Qdrant~\citep{qdrant} and LanceDB~\citep{lancedb} are vector databases
built in Rust using HNSW for approximate search;
and kiddo~\citep{kiddo} and hora~\citep{hora} provide KD-tree and
approximate nearest neighbor search, respectively.
These projects address approximate search or coordinate-specific structures.
\rustknn{} provides \emph{exact} search over \emph{arbitrary} metric spaces
with the full algorithmic suite from~\citet{izbicki2015faster}, filling a gap
in the Rust ML toolchain.

%% ===========================================================================
\section{Background}
\label{sec:background}
%% ===========================================================================

\subsection{Metric Spaces and the Doubling Dimension}

A \emph{metric space} $(\mathcal{X}, d)$ is a set $\mathcal{X}$ equipped with a
distance function $d : \mathcal{X} \times \mathcal{X} \to \mathbb{R}_{\ge 0}$
satisfying non-negativity, identity of indiscernibles ($d(x,y) = 0 \iff x = y$),
symmetry, and the triangle inequality.  Given a finite dataset
$S \subset \mathcal{X}$, the \emph{doubling constant}~$c$ is the minimum value
such that every ball of radius~$r$ centered on a point in~$S$ can be covered by
at most~$c$ balls of radius~$r/2$~\citep{krauthgamer2004navigating,
clarkson1999nearest}.  The \emph{doubling dimension} is
$\dim_D = \log_2 c$.  Crucially, $c$~is a property of the dataset, not of the
ambient space: data lying on a low-dimensional manifold embedded in
$\mathbb{R}^D$ can have small~$c$ even when $D$~is large.  Cover tree query
complexity is $O(c^6 \log n)$, parametrized by~$c$ rather than~$D$---this is
what makes cover trees effective in high dimensions, unlike KD-trees whose
performance degrades exponentially with~$D$.

\subsection{Cover Tree Invariants}

A cover tree is a leveled tree in which each node stores a single data point.
Intuitively, each level represents a different \emph{resolution}: the root sees
the dataset at coarse scale, and deeper nodes refine the picture.  Three
structural invariants govern parent-child relationships
(Definition~1 of~\citet{beygelzimer2006cover}):

\begin{enumerate}
    \item \textbf{Leveling invariant.} Every node~$p$ has an integer
    $\mathrm{level}(p)$, and every child~$q$ satisfies
    $\mathrm{level}(q) = \mathrm{level}(p) - 1$.
    Each level down is a finer resolution.

    \item \textbf{Covering invariant.} Define the covering distance
    $\covdist(p) = b^{\mathrm{level}(p)}$.  For each child~$q$ of~$p$,
    $d(p,q) \le \covdist(p)$.
    No child lies beyond its parent's covering radius.

    \item \textbf{Separating invariant.} Define the separation distance
    $\sepdist(p) = b^{\mathrm{level}(p)-1}$.  For distinct children $q_1, q_2$
    of~$p$, $d(q_1, q_2) > \sepdist(p)$.
    No two siblings are too close together.
\end{enumerate}

\noindent
The base parameter~$b$ controls the ratio between covering and separating
distances.  The original formulation uses $b = 2$, but
\citet{beygelzimer2006cover} found empirically that $b = 1.3$ reduces distance
computations by 10--71\% on benchmark datasets.  Our experiments default to
$b = 1.3$.

Each node also caches
$\maxdist(p) = \max_{q \in \mathrm{descendants}(p)} d(p, q)$, the greatest
distance from~$p$ to any of its descendants, which is upper bounded by
$\covdist(p) \cdot b/(b{-}1)$ (a geometric series over descendant levels;
this equals $2 \cdot \covdist(p)$ when $b{=}2$ and ${\approx}4.3 \cdot
\covdist(p)$ at the default $b{=}1.3$).  This value enables the key pruning rule during queries:
a subtree rooted at~$q$ can be skipped entirely when
$d(\mathrm{query}, q) - \maxdist(q) > d_{\mathrm{best}}$, because even the
closest descendant of~$q$ must be farther than the current best candidate.

From \citet{beygelzimer2006cover}, the maximum number of children per node is
$O(c^4)$ and the maximum tree depth is $O(c^2 \log n)$.

\subsection{Algorithmic Variants}

The original cover tree~\citep{beygelzimer2006cover} was defined as an implicit
infinite structure in which every internal node also appears as its own child---the
\emph{nesting invariant}.  The explicit representation stores $O(n)$ nodes.
\citet{izbicki2015faster} observed that the nesting invariant is unnecessary for
the runtime analysis and introduced two improved variants that eliminate
self-children, reducing the node count to exactly~$n$.

\paragraph{Simplified cover tree construction.}
The simplified insertion of~\citet{izbicki2015faster} builds a tree via
depth-first search: starting from the root, the algorithm finds the first child
whose covering distance accommodates the new point and recurses; if no child
qualifies, the point becomes a new child at $\mathrm{level} - 1$.  If the new
point lies beyond the root's covering distance, the root level is raised. The tree
contains exactly~$n$ nodes, one per data point, eliminating the self-children of
the original formulation.  Construction runs in $O(c^6 n \log n)$ time.

\paragraph{Nearest ancestor cover tree construction.}
This variant adds a fourth invariant: for every point~$p$ with ancestor~$q_1$ and
sibling~$q_2$ of~$q_1$, we require $d(p, q_1) \le d(p, q_2)$---every point's
ancestor at each level is the closest possible.  Insertion sorts children by
distance to the new point and invokes a \emph{rebalance} procedure that extracts
descendants violating the invariant and reinserts them under the new point.
Construction is costlier, but the resulting tree has tighter $\maxdist$~values,
leading to more aggressive pruning; the trees are also deeper, so querying is not
always quicker than the simplified variant.

\paragraph{Parallel cover tree construction.}
Two cover trees are merged by matching children between them at each level,
enabling parallel construction: partition the dataset across processors, build
independent trees, then merge via binary reduction in $O(\log p)$ rounds.
Roughly 1\% of distance computations occur during the merge; the bottleneck is
cache performance in the parallel construction phase. We implement this but do
not time it in our experiments; Table~\ref{tab:app_dist_full} reports
its distance counts.

\paragraph{Single-tree query.}
Nearest-neighbor search traverses the tree depth-first from the root,
visiting children in order of increasing distance to the query.
A child~$q$ is pruned when
$d(\mathrm{query}, q) - \maxdist(q) > d_{\mathrm{best}}$, yielding
$O(c^6 \log n)$ runtime.

\paragraph{Batch single-tree query.}
This algorithm, derived from the code of~\citet{beygelzimer2006cover} but not
described in detail in their paper, organizes the query set into its own cover
tree and maintains \emph{cover sets}---lists of candidate reference nodes,
indexed by scale level---that are passed top-down through the query tree
hierarchy.  At each scale level, three cases arise
(Figure~\ref{fig:batch_st} in the appendix):
(i)~\emph{reference descent}---expand each candidate reference node by replacing
it with its children, computing distances and pruning;
(ii)~\emph{query split}---when a query node has children at the current scale,
each query child receives a \emph{filtered copy} of the parent's cover sets with
distances recomputed and entries pruned against the child's bound; the parent
point continues as a ``virtual self-child,'' inheriting the cover sets directly;
(iii)~\emph{base case}---at the leaf scale, score remaining candidates.
The key mechanism is \emph{cover-set inheritance}: once a reference candidate is
pruned from a query node's cover set, no descendant query point will ever see it.
A mutable upper bound threaded through the recursion means any improvement from
scoring one query point immediately tightens the threshold for siblings and
descendants.  The amortization scales with query tree depth: in all-NN mode
(query~=~reference) the query tree has full depth $O(c^2 \log n)$ with many split
points, maximizing amortization, whereas a small held-out query set yields only
2--3 levels, reducing most queries to independent base-case evaluations.

\paragraph{Dual-tree traversal.}
Dual-tree traversal~\citep{ram2009linear,curtin2013tree} simultaneously
descends query and reference trees, pruning entire (query-subtree,
reference-subtree) pairs via the triangle inequality---a joint pruning
that can discard more work per distance computation than batch single-tree's
per-query bounds, at the cost of more bookkeeping per node pair. We summarize this algorithm in Figure~\ref{fig:dual_tree} in the Appendix. 

\paragraph{Why interactions matter.}
These three query modes, combined with the base parameter~$b$ and
micro-optimizations, create a
multi-dimensional performance landscape.
As we show in Section~\ref{sec:experiments}, the optimal configuration
depends on~$k$, dataset geometry, and evaluation protocol in ways that
are not apparent from any single comparison.

%% ===========================================================================
\section{Implementation and Optimization Ablation}
\label{sec:implementation}
%% ===========================================================================

\subsection{Implementation Overview}

\rustknn{} comprises approximately 13{,}000 lines of Rust organized around
a generic \texttt{Distance<T>} trait that enables arbitrary metric spaces.
After cover tree construction, pointer-based trees are converted to a contiguous
\texttt{PackedCoverTree} in depth-first order (a simpler alternative to the van Emde Boas layout of~\citet{izbicki2015faster})
for cache locality: each packed node occupies ${\sim}$40 bytes, fitting within a single 64-byte cache line.
The dual-tree traversal adapts the framework of~\citet{curtin2013tree}
to cover trees without self-children, using a three-phase expansion with
triangle inequality pre-filters (details in
Appendix~\ref{sec:app_impl}).

\subsection{Optimization Ablation}
\label{sec:ablation}

Five micro-optimizations are implemented as compile-time feature flags,
enabling controlled ablation experiments.\footnote{A sixth optimization---sorted-vector batch tracking---replaces a
hash map with a sorted vector for mapping held-out batch results back to their inputs.
Because the ablation infrastructure targets the all-NN benchmark in
dual-tree mode (\rustknn{} Packed), this batch-only optimization is not evaluated;
its impact is expected to scale with the number of concurrent query
points in batch mode. }
These affect query algorithms and are evaluated in
Table~\ref{tab:ablation}:
\begin{enumerate}
  \item \textbf{Early-exit distance}: Euclidean distance computation
    with four independent accumulators and periodic bound checks; exits early when the partial sum exceeds the pruning bound, avoiding
    the full $O(d)$ computation.
  \item \textbf{Halfsort}: Partial sort of children,
    $O(n)$ versus $O(n \log n)$ full sort.
  \item \textbf{SmallVec}: Stack-allocated child arrays for the typical
    3--5 children per node at base~1.3.
  \item \textbf{$k{=}1$ fast path}: Dedicated single-comparison path
    in the $k$-NN state machine.
  \item \textbf{Triangle pre-filters}: Each reference child caches its parent
    distance $d_{\mathrm{parent}}(r_c)$ at construction time.  Before computing
    $d(q, r_c)$, the triangle inequality gives a free lower bound
    $|d(q, r_p) - d_{\mathrm{parent}}(r_c)|$ using the already-known parent
    score $d(q, r_p)$.  If this lower bound minus the child's $\maxdist$
    exceeds the current $k$th-best distance, the entire subtree is skipped
    without a metric evaluation.
\end{enumerate}

Table~\ref{tab:ablation} shows the slowdown factor when each optimization
is individually disabled, measured at $k{=}1$ (all-NN benchmark) on five
datasets chosen to span the dimensionality axis (pima $d{=}8$, pendigits
$d{=}16$, corel $d{=}32$, covtype $d{=}54$, mnist $d{=}784$), since the
headline question here---whether an optimization's benefit scales with~$d$---is
a claim about that axis. We exclude the five smallest UCI sets ($n \le 351$):
their millisecond-scale runtimes are dominated by measurement noise, as the
intervals below make concrete. Each cell is the median over
\textbf{20 independent SLURM allocations}, with a bootstrap 95\% confidence
interval; Section~\ref{sec:experiments} (Methodology) explains why we report
between-allocation rather than within-allocation spread.
The ``all off'' row disables all five optimizations simultaneously. Full per-dataset, per-$k$ ablation tables appear in
Appendix~\ref{sec:app_ablation}.

\begin{table}[!htb]
  \caption{Ablation: slowdown ($\times$ baseline) when each optimization
    is disabled ($k{=}1$, all-NN, base~1.3).
    Median [95\% CI] over 20 independent allocations.
    Values ${>}1$ mean the optimization helps; intervals spanning~1.0
    indicate no reliable effect.
    Computed from the same pooled sample as the waterfall in
    Table~\ref{tab:app_waterfall}, so the ``all off'' row here and that
    table's ``baseline'' row are identical by construction.}
  \label{tab:ablation}
  \centering
  \footnotesize
  \setlength{\tabcolsep}{3pt}
  \begin{tabular}{l ccccc}
    \toprule
    Disabled optimization & pima & pendigits & corel & covtype & mnist \\
                          & ($d{=}8$) & ($d{=}16$) & ($d{=}32$) & ($d{=}54$) & ($d{=}784$) \\
    \midrule
    Early-exit distance    & 0.89 [0.73, 1.24] & 0.82 [0.63, 0.99] & 0.98 [0.97, 1.00] & 1.31 [1.27, 1.36] & 1.49 [1.39, 1.73] \\
    Halfsort               & 1.28 [1.02, 1.42] & 0.96 [0.75, 1.10] & 1.10 [1.09, 1.11] & 1.03 [1.03, 1.04] & 1.02 [1.02, 1.03] \\
    SmallVec               & 0.97 [0.89, 1.09] & 0.65 [0.61, 0.95] & 0.99 [0.98, 1.04] & 1.01 [1.00, 1.02] & 1.02 [1.01, 1.03] \\
    $k{=}1$ fast path      & 1.06 [0.95, 1.28] & 0.78 [0.69, 0.99] & 1.04 [0.97, 1.06] & 1.03 [1.03, 1.04] & 1.01 [1.01, 1.02] \\
    Triangle pre-filter    & 1.40 [1.28, 1.52] & 0.75 [0.71, 0.79] & 1.01 [0.98, 1.03] & 0.98 [0.97, 0.99] & 1.03 [0.96, 1.04] \\
    All off                & 1.11 [0.99, 1.31] & 0.78 [0.72, 0.83] & 1.06 [1.03, 1.07] & 1.33 [1.31, 1.34] & 1.58 [1.39, 1.60] \\
    \bottomrule
  \end{tabular}
\end{table}

Early-exit distance is the one optimization with a consistent effect, and
its benefit grows with dimension: $1.31\times$ on covtype ($d{=}54$) and
$1.49\times$ on mnist ($d{=}784$), but none on corel ($d{=}32$, CI
$[0.97, 1.00]$). The effect is purely per-call work: disabling early exit
leaves the dual-tree distance count unchanged on every ablation dataset
(159.6M calls on corel either way), so the traversal is identical and only
the cost of each rejected candidate differs. In fact none of the five flags
changes the count by more than 0.1\%; their time effects are constant-factor
effects.

The confidence intervals make the noise floor explicit, which is essential to
reading the table correctly. On corel, covtype, and mnist they are tight, so
the effects there are measurable. On pima and pendigits they are not: their
per-allocation timings are \emph{bimodal}---on pendigits, allocations
of the same binary cluster near 170\,ms or near 260\,ms, and the same split
appears in an independent earlier sweep---so a ratio of two medians reflects
which mode each binary's allocations happened to land in rather than the
optimization. The sub-1.0 entries in those columns (e.g., pendigits SmallVec
$0.65\times$) are of this kind. Conclusions about micro-optimizations on
small datasets are therefore not supportable, and we draw none.

Among the other optimizations, none has a consistent effect: halfsort,
SmallVec, the $k{=}1$ fast path, and the triangle pre-filter are within
4\% of $1.0\times$ on covtype and mnist, and on corel only halfsort
($1.10\times$) exceeds that. Differences of this size
also arise between builds that should behave identically---for example,
disabling the $k{=}1$ fast path makes corel's $k{\ge}5$ queries 6--9\% faster
(Table~\ref{tab:app_ablation_perk}), although that path is never taken at
$k{>}1$---so we treat effects of up to about 10\%, corel's halfsort
included, as code-layout noise rather than as properties of the optimization.

%% ===========================================================================
\section{Experiments}
\label{sec:experiments}
%% ===========================================================================

\subsection{Experimental Setup}
We describe the hardware and software setup for our experiments in Appendix~\ref{sec:exp_impl}.

\paragraph{Datasets.}
Table~\ref{tab:datasets} lists the 12~benchmark datasets.
Eleven originate from the UCI Machine Learning
Repository~\citep{dua2019uci}; MNIST is from the original LeCun
distribution~\citep{lecun1998mnist}.
Sizes span three orders of magnitude ($n = 150$ to $581{,}012$) and
dimensionalities range from~4 to~784, providing broad coverage of
the regimes where cover trees are applicable.
The smallest sets are included for two reasons: coverage of the $(n,d)$ plane,
and direct comparability with the seminal cover tree
evaluations~\citep{beygelzimer2006cover, izbicki2015faster}, which use exactly
these sets---re-evaluating that literature under controlled conditions is part
of this paper's purpose. They are not the high-performance target, and as
Section~\ref{sec:ablation} quantifies, their millisecond-scale timings are
noise-dominated; the performance conclusions rest on corel, covtype, and mnist.
Appendix~\ref{sec:app_synthetic} adds non-tiny synthetic data ($n{=}20{,}000$)
whose intrinsic dimension is known by construction, which real datasets cannot
provide.
All experiments use the Euclidean ($L_2$) distance.

\begin{table}
  \vspace{-1em}
  \caption{Benchmark datasets used in all experiments.
    $n$ is the number of points and $d$ is the dimensionality.}
  \label{tab:datasets}
  \centering
  \small
  \begin{tabular}{lrrl @{\hspace{2em}} lrrl}
    \toprule
    Dataset & $n$ & $d$ & Source & Dataset & $n$ & $d$ & Source \\
    \midrule
    iris        &   150 &   4 & UCI & optdigits &   5{,}620 &  64 & UCI \\
    wine        &   178 &  13 & UCI & pendigits &  10{,}992 &  16 & UCI \\
    glass       &   214 &   9 & UCI & letter    &  20{,}000 &  16 & UCI \\
    bupa        &   345 &   6 & UCI & corel     &  37{,}749 &  32 & UCI \\
    ionosphere  &   351 &  34 & UCI & mnist     &  60{,}000 & 784 & LeCun \\
    pima        &   768 &   8 & UCI & covtype   & 581{,}012 &  54 & UCI \\
    \bottomrule
  \end{tabular}
  \vspace{-1em}
\end{table}

\paragraph{Methodology.}
Every configuration (dataset $\times$ algorithm $\times$ variant $\times$
query mode $\times$ $k$) is executed as \textbf{20 independent SLURM
allocations}---20 separately scheduled jobs, not 20 repetitions inside one job.
We reduce each allocation to one number (the median of its internal timed runs)
and report the \emph{median over the 20 allocations} with a bootstrap 95\%
confidence interval (4{,}000 resamples, seeded). 

This design captures a distinction that matters more than we initially
expected: repeated runs within one allocation agree to 1--2\%, but
two allocations of a byte-identical binary can differ by 16--29\% on
small, fast datasets;
our intervals report the between-allocation component.
Section~\ref{sec:ablation} shows a case where this is decisive.
Wall-clock time is measured with \texttt{std::time::Instant} (Rust) and
\texttt{std::chrono::high\_resolution\_clock} (C++).
Construction and query times are reported separately; query time is
further decomposed into tree-build, traversal, and result-collection
phases where applicable.
Distance computation counts are obtained from a separately instrumented
binary (compiled with \texttt{-{}-features instrument}), which increments
a thread-local counter on every call to the distance function---including
early-exited calls that return before computing all dimensions. Because
the counter counts \emph{invocations}, not full $O(d)$ evaluations,
early-exit distance reduces the per-call \emph{work} but not the count; in the
dual-tree traversal it leaves the number of invocations unchanged
(Section~\ref{sec:ablation}).
Table~\ref{tab:app_dist_full} reports counts for the dual-tree all-NN mode,
the mode the ablation (Section~\ref{sec:ablation}) measures.
Results were collected in several measurement \emph{campaigns} (separately
queued batches of allocations); between campaigns, medians can shift by
20--40\% due to node placement. Every ranking or ratio from which we draw a conclusion is computed
within a single table and therefore within a single campaign.
Absolute large-dataset times also depend on node load: on an idle node, an mnist
all-NN slows $1.4$--$1.5\times$ when 15 copies share the node and $2.0$--$2.1\times$
with 60, about equally for \rustknn{} and C++. All campaigns therefore ran at a
similar, high load, so compared implementations saw comparable conditions
(Appendix~\ref{sec:exp_impl}).
\paragraph{Baselines.}
We compare against three established baselines:
\begin{enumerate}
  \item \textbf{C++ reference}:
    the original cover tree implementation of~\citet{beygelzimer2006cover},
    adapted to produce JSON output; its batch all-nearest-neighbor query (from
    which \rustknn{}'s batch single-tree mode derives).
  \item \textbf{mlpack single-tree}:
    mlpack~4.7.1's cover tree with single-tree $k$-NN search and base~1.3.
  \item \textbf{mlpack dual-tree}:
    mlpack~4.7.1's cover tree with dual-tree traversal and base~1.3.
\end{enumerate}
All baselines consume the same CSV data files and use Euclidean
distance.
These three baselines are compared against \rustknn{} across all
12~datasets.

\subsection{Query Mode $\times$ $k$ Interaction}
\label{sec:mode_k}

Table~\ref{tab:allnn} presents the all-nearest-neighbor benchmark ($k{=}1$):
for each point in the dataset, find its nearest neighbor.
This is the standard cover tree benchmark
from~\citet{izbicki2015faster}, exercising tree construction and exhaustive
querying.  \rustknn{} results use packed
simplified cover trees in three query modes (dual-tree, single-tree, batch
single-tree).

\begin{table}[!h]
  \caption{All-nearest-neighbor query time (ms, $k{=}1$, base~1.3).
    Median [95\% CI] over 20 independent SLURM allocations.
    \rustknn{} results show three query modes: dual-tree, single-tree,
    and batch single-tree. Best result per row in \textbf{bold}.
    $k{=}10$ results in Appendix Table~\ref{tab:allnn_k10}.
    Values ${\ge}10^3$ are abbreviated K/M.}
  \label{tab:allnn}
  \centering
  \scriptsize
  \setlength{\tabcolsep}{1.5pt}
  \begin{tabular}{l@{\hspace{3pt}}c rr r rrr}
    \toprule
    & & \multicolumn{3}{c}{\rustknn{}} & C++ & mlpack & mlpack \\
    \cmidrule(lr){3-5}
    Dataset & $d$ & Dual & Single & Batch & Ref & Single & Dual \\
    \midrule
    iris & 4 & 0.82 [0.80, 0.84] & 0.37 [0.36, 0.38] & 0.71 [0.68, 0.72] & \textbf{0.33 [0.28, 0.35]} & 1.24 [1.03, 1.30] & 1.32 [1.16, 1.35] \\
    wine & 13 & 0.82 [0.74, 0.85] & 0.66 [0.57, 0.68] & 0.74 [0.66, 0.76] & \textbf{0.35 [0.27, 0.35]} & 1.18 [0.94, 1.20] & 1.01 [0.81, 1.02] \\
    glass & 9 & 1.77 [1.41, 1.86] & 0.96 [0.83, 1.05] & 1.36 [1.15, 1.50] & \textbf{0.54 [0.48, 0.60]} & 2.51 [2.23, 2.75] & 2.22 [2.02, 2.46] \\
    bupa & 6 & 4.02 [3.48, 4.21] & 2.07 [1.77, 2.21] & 3.13 [2.58, 3.23] & \textbf{1.27 [1.18, 1.41]} & 6.02 [5.64, 6.47] & 6.17 [5.60, 6.77] \\
    pima & 8 & 7.29 [6.62, 8.66] & 9.05 [8.09, 10.8] & 6.68 [5.67, 7.38] & \textbf{2.43 [2.26, 2.50]} & 13.5 [12.5, 13.9] & 13.2 [11.9, 13.3] \\
    ionosphere & 34 & 4.50 [4.25, 4.76] & 2.54 [2.44, 2.68] & 4.13 [3.91, 4.38] & \textbf{1.16 [1.09, 1.19]} & 6.92 [6.59, 7.23] & 6.61 [6.41, 6.98] \\
    optdigits & 64 & 887 [886, 888] & \textbf{160 [159, 165]} & 979 [973, 1{,}129] & 264 [260, 284] & 1.23K [1.2, 1.31] & 1.68K [1.66, 1.79] \\
    pendigits & 16 & 174 [173, 191] & 144 [143, 153] & 230 [228, 242] & \textbf{96.6 [87.1, 97.9]} & 560 [546, 601] & 584 [583, 618] \\
    letter & 16 & 1.13K [1.12, 1.18] & 644 [640, 680] & 2.02K [1.98, 2.15] & \textbf{576 [568, 605]} & 3.78K [3.68, 3.9] & 6.44K [6.17, 6.6] \\
    corel & 32 & 4.96K [4.88, 5.17] & \textbf{2.93K [2.69, 2.96]} & 17.2K [17.1, 17.4] & 4.69K [4.31, 4.86] & 25.1K [20.9, 25.2] & 52K [40.7, 52.2] \\
    covtype & 54 & 13.6K [12, 14.3] & 102K [88.8, 103] & 54.8K [45.6, 57.7] & \textbf{13K [11.8, 13.2]} & 68.9K [62, 70.3] & 80.9K [72.6, 89] \\
    mnist & 784 & 673K [584, 705] & \textbf{381K [307, 460]} & 2.34M [2.06, 3.09] & 1.40M [1.15, 1.83] & 4.86M [3.85, 5.89] & 3.85M [3.25, 4.54] \\
    \bottomrule
  \end{tabular}
\end{table}

The key finding is that no single implementation or query mode dominates.
At $k{=}1$, the C++ reference has the lowest median on 9 of 12~datasets---every
dataset with $n \le 20{,}000$ except optdigits, plus covtype, where its interval
overlaps \rustknn{} Dual---while \rustknn{} single-tree search wins on optdigits,
corel, and mnist, by $1.7\times$, $1.6\times$, and $3.7\times$ over the C++ reference.
At $k{=}10$ (Appendix Table~\ref{tab:allnn_k10}) the C++ reference still wins
every dataset with $n \le 20{,}000$, but \rustknn{} dual-tree search wins the
three largest, by $1.9\times$ on corel, $2.2\times$ on covtype, and $3.6\times$
on mnist over the C++ reference.

\paragraph{$k$-scaling dynamics.}
Table~\ref{tab:k_scaling} traces the best \rustknn{} mode across five values
of~$k$ for four representative datasets.
The crossover is sharpest on mnist: single-tree search is $1.8\times$ faster
than dual-tree search at $k{=}1$ but $3.1$--$3.5\times$ slower at every
$k{\ge}5$. Corel shows the same pattern at smaller scale (single-tree
$1.7\times$ faster at $k{=}1$, dual-tree $1.2$--$1.6\times$ faster at
$k{\ge}5$); on optdigits single-tree leads through $k{=}10$, the two tie at
$k{=}25$, and dual-tree leads at $k{=}50$. On covtype dual-tree search is best
at every~$k$, and the largest mode-choice penalty in the table is there:
single-tree search is $7.0$--$7.5\times$ slower at every~$k$.
Batch single-tree is never the fastest mode on these four datasets; at
$k{\ge}5$ it is within 30\% of the best only on optdigits.
On mnist and covtype the gaps are well outside the confidence intervals; on
corel they overlap slightly at $k{=}5$, and on optdigits the modes are within
their intervals throughout.

\begin{table}[!h]
  \caption{Query mode $\times$ $k$ interaction (ms, all-NN, \rustknn{} Packed).
    Median [95\% CI] over 20 independent allocations. Bold = best mode per
    $(\text{dataset}, k)$. Values ${\ge}10^3$ are abbreviated K/M.}
  \label{tab:k_scaling}
  \centering
  \scriptsize
  \setlength{\tabcolsep}{2pt}
  \begin{tabular}{ll rrrrr}
    \toprule
    Dataset & Mode & $k{=}1$ & $k{=}5$ & $k{=}10$ & $k{=}25$ & $k{=}50$ \\
    \midrule
    \multirow{3}{*}{optdigits}
      & Dual & 887 [886, 888] & 1.09K [1.08, 1.1] & 935 [837, 1{,}184] & 979 [963, 1{,}106] & \textbf{1.13K [1.12, 1.2]} \\
      & Single & \textbf{160 [159, 165]} & \textbf{933 [885, 1{,}156]} & \textbf{772 [736, 902]} & \textbf{976 [962, 1{,}077]} & 1.21K [1.2, 1.26] \\
      & Batch & 979 [973, 1{,}129] & 1.16K [1.06, 1.19] & 987 [969, 1{,}143] & 1.14K [1.13, 1.21] & 1.34K [1.32, 1.42] \\
    \midrule
    \multirow{3}{*}{corel}
      & Dual & 4.96K [4.88, 5.17] & \textbf{6.55K [6.3, 6.84]} & \textbf{7.58K [7.27, 7.86]} & \textbf{9.42K [8.97, 9.75]} & \textbf{11.5K [10.8, 11.9]} \\
      & Single & \textbf{2.93K [2.69, 2.96]} & 8.12K [6.78, 8.32] & 9.88K [8.22, 10.2] & 13.1K [10.8, 13.5] & 18.1K [13.5, 18.4] \\
      & Batch & 17.2K [17.1, 17.4] & 21.4K [16.9, 21.5] & 23.6K [19, 24] & 28.1K [22, 28.9] & 32.8K [25.5, 34.1] \\
    \midrule
    \multirow{3}{*}{covtype}
      & Dual & \textbf{13.6K [12, 14.3]} & \textbf{20.4K [17.7, 21]} & \textbf{25.2K [22.1, 25.7]} & \textbf{37K [32.9, 37.2]} & \textbf{53.2K [47.9, 53.4]} \\
      & Single & 102K [88.8, 103] & 151K [131, 154] & 185K [160, 190] & 262K [226, 273] & 374K [315, 387] \\
      & Batch & 54.8K [45.6, 57.7] & 91.7K [77, 94.4] & 121K [103, 123] & 182K [153, 182] & 252K [209, 254] \\
    \midrule
    \multirow{3}{*}{mnist}
      & Dual & 673K [584, 705] & \textbf{811K [741, 864]} & \textbf{886K [766, 950]} & \textbf{956K [890, 1{,}073]} & \textbf{1.07M [0.94, 1.16]} \\
      & Single & \textbf{381K [307, 460]} & 2.49M [2.46, 3.60] & 2.90M [2.62, 3.96] & 3.38M [2.90, 3.99] & 3.49M [2.83, 4.63] \\
      & Batch & 2.34M [2.06, 3.09] & 2.52M [2.42, 3.53] & 2.98M [2.70, 3.76] & 3.12M [2.66, 4.07] & 3.63M [2.88, 4.61] \\
    \bottomrule
  \end{tabular}
\end{table}

\paragraph{Why the best mode changes with $k$.}
Single-tree search runs one independent descent per query, pruning against
that query's current $k$th-best distance. At $k{=}1$ that radius is the
nearest-neighbor distance, which is small, and the dedicated $k{=}1$ path
keeps the per-candidate bookkeeping to a single comparison: on mnist the
single-tree all-NN issues $6\times$ fewer distance computations than the
dual-tree traversal (312M vs.\ 1.87B, Table~\ref{tab:app_dist_full}).
As $k$ grows, the pruning radius widens to the $k$th-nearest distance and
every one of the~$n$ independent descents pays for it separately: from
$k{=}1$ to $k{=}5$, single-tree time on mnist grows $6.5\times$ while
dual-tree time grows $1.2\times$ (Table~\ref{tab:k_scaling}).

Dual-tree traversal~\citep{curtin2013tree} amortizes that cost by pruning
entire (query-subtree, reference-subtree) pairs, but it pays a per-pair
overhead in exchange. When both nodes are internal it runs a three-phase
expansion, producing $O(|\mathrm{children}_q| \times |\mathrm{children}_r|)$
candidate pairs per internal node---typically 9--25 at base~1.3, each a
recursive call---and it maintains a per-query-node bound cache with epoch
invalidation, and computes phase-1/2 distances that are wasted whenever the
pairs they would pre-filter are pruned anyway. Where pruning is weak relative
to that overhead---small datasets, or $k{=}1$, where single-tree search
already prunes tightly---the overhead dominates; on the three largest
datasets at $k{\ge}5$, and on covtype at every~$k$, joint pruning wins.
Batch single-tree search, which shares reference expansion across a query
subtree but scores each query separately, gets neither benefit here: it is
the slowest of the three modes on corel and optdigits at every~$k$ and on mnist
at four of five, and lands between the other two only on covtype and on mnist at $k{=}25$.
Appendix~\ref{sec:app_impl} and Figures~\ref{fig:batch_st}--\ref{fig:dual_tree}
give the algorithms and worked diagrams.

The held-out results (Table~\ref{tab:heldout}) show a different pattern:
with a shallow query tree ($n_q \le 512$), dual-tree's construction overhead is
reduced but so is its pruning leverage.

\subsection{Base Parameter Analysis}
\label{sec:base_analysis}

\citet{beygelzimer2006cover} recommend base~$b{=}1.3$ based on empirical
evaluation. For \rustknn{}'s dual-tree all-NN (Table~\ref{tab:optimal_base},
Appendix~\ref{sec:app_base}), our data do not support it as a default.

Three patterns emerge.
First, \textbf{coarser bases win}: the optimum is $b{=}2.0$---the value used in
the original analysis---on four of the seven datasets in
Table~\ref{tab:optimal_base} at $k{=}1$ and five at $k{=}10$, and $b{=}3.0$ on
the two smallest (glass and ionosphere, $n \le 351$), where shallower trees
reduce the traversal overhead that dominates at small scale. Across all twelve
datasets (Appendix~\ref{sec:app_base}), $b{=}1.3$ is optimal in only one of
24 (dataset, $k$) cases.
Second, \textbf{the optimum rarely shifts with $k$}: in
Table~\ref{tab:optimal_base} only mnist moves ($1.3$ at $k{=}1$, $2.0$ at
$k{=}10$), and there the optimum is shallow: base~2.0 is 10\% slower than 1.3
at $k{=}1$ and within 1\% of it at $k{=}10$.
Third, \textbf{the default's cost is real where it matters}: $b{=}1.3$ is
$1.57\times$ slower than the optimum on covtype at $k{=}1$ ($1.51\times$ at
$k{=}10$) and $1.11$--$1.17\times$ slower on corel and pendigits; it is within
2\% of the optimum only on optdigits at $k{=}1$ and on mnist.
We therefore recommend sweeping the base rather than adopting 1.3 unchecked:
a sweep over $b \in \{1.3, 2.0, 3.0\}$ costs seconds on most datasets (roughly
10--15 minutes per base on mnist). We do not generalize this beyond dual-tree
search; the per-implementation curves in Appendix~\ref{sec:app_base} show how
the other implementations respond to the base.
Full base sensitivity curves appear in Appendix~\ref{sec:app_base}.

\subsection{All-NN vs.\ Held-Out Divergence}
\label{sec:heldout_divergence}

The all-nearest-neighbor benchmark, in which query and reference sets
coincide, is the standard evaluation in KNN
literature~\citep{beygelzimer2006cover, izbicki2015faster}. This setting can enable some memory and algorithmic optimizations since the query and reference sets can use the same cover tree. However, most practical workloads involve  held-out queries against a
pre-built index.
Table~\ref{tab:heldout} (Appendix~\ref{sec:app_heldout_a}) shows held-out
performance diverges from all-NN rankings.

\paragraph{Ranking divergence.}
The clearest case is mlpack's two query modes. In the all-NN benchmark they are
within about $2\times$ of each other on every dataset, and dual-tree is the
faster of the two on five of them (Table~\ref{tab:allnn}); held out, mlpack-Dual is $7.8\times$ slower
than mlpack-Single on pendigits and $250\times$ slower on covtype at
$n_q{=}512$ (Table~\ref{tab:heldout}). An evaluation that used only the
all-NN benchmark would never see a gap of that size.
The best \rustknn{} mode changes in the same way: on mnist, single-tree search
wins the all-NN benchmark at $k{=}1$, but held out at $n_q{=}512$ dual-tree
search is $1.8\times$ faster than single-tree (8.4 vs.\ 15.1\,s;
Appendix~\ref{sec:app_heldout_a}). And on covtype, the C++ reference ties
\rustknn{}'s best mode in all-NN at $k{=}1$ (13.0K vs.\ 13.6K\,ms, intervals
overlapping) but is $1.5\times$ faster held out at $n_q{=}512$ (53.0 vs.\
81.1\,ms, intervals disjoint).
This divergence is systematic: in all-NN mode the query tree is the reference
tree, so dual-tree and batch methods amortize their tree-building and
expansion costs over all $n$ points; held out, the query tree is shallow and
that amortization is small.

\paragraph{Where our implementation loses.}
Table~\ref{tab:heldout} also contains a result worth stating plainly rather than
glossing: on covtype held-out---arguably the most practically relevant cell in
the paper, being the one dataset with both large~$n$ and moderate~$d$---the
C++ reference is fastest at every $n_q$, and \rustknn{}'s best mode (dual-tree)
is $1.2$--$1.5\times$ slower. \rustknn{}-Single is also slower than
mlpack-Single at $n_q \ge 128$ (176 vs.\ 142\,ms at $n_q{=}512$). But
single-tree is the wrong \rustknn{} mode here: \rustknn{}-Dual is $2.2\times$
faster than \rustknn{}-Single at $n_q{=}512$ and $1.75\times$ faster than
mlpack-Single. Both facts are the same lesson from opposite directions---an
implementation evaluated in only one query mode gets misranked, whichever way
the misranking happens to fall.

\paragraph{Implications for evaluation.}
We recommend reporting both all-NN and held-out results: the former stresses
scalability, the latter approximates serving latency.
Full held-out scaling curves appear in
Appendices~\ref{sec:app_heldout_a}--\ref{sec:app_heldout_b}.

\subsection{Beyond Cover Trees: A KD-Tree Under the Same Harness}
\label{sec:beyond_ct}

The factors we isolate above are cover-tree-specific in their details, which
raises the question of whether the \emph{methodological} finding---that
implementation choices can dominate algorithmic ones---holds for other index
structures. To test this without rewriting the study, we added a median-split
KD-tree \emph{inside \rustknn{}}, reusing the identical kernel,
candidate structure, distance counter, and timing code, so it runs the same five feature-flag
ablations. We also benchmarked \texttt{kiddo}, an independent Rust KD-tree crate,
under the same 20-allocation protocol, separating the effect of the codebase from
that of the index. Full tables are in Appendix~\ref{sec:app_kdtree}.

\textbf{The better index reverses with dimension.} On all-NN at $k{=}1$, the
KD-tree is $13\times$ faster than single-tree cover trees on corel
(254 vs.\ 3{,}248\,ms; same-allocation values from Table~\ref{tab:kdtree}) and
$23\times$ faster on covtype (4{,}465 vs.\ 100{,}590\,ms), because axis-aligned
box pruning is highly effective at moderate~$d$. At $d{=}784$ that pruning
collapses---the KD-tree issues 594M distance
computations---and it becomes $12\times$ \emph{slower} than the single-tree
cover tree (4.09M vs.\ 334{,}487\,ms): a textbook curse-of-dimensionality
crossover, measured with no cross-codebase confounders.

\textbf{The same algorithm in a different codebase differs by up to
$9\times$.} Our KD-tree and \texttt{kiddo} implement the same algorithm, yet
\texttt{kiddo} is $9\times$ faster on pendigits (7.8 vs.\ 72\,ms) while being
$2.9\times$ \emph{slower} on covtype and $2.0\times$ slower on mnist (8.31M
vs.\ 4.09M\,ms): each wins on different data. A practitioner comparing ``KD-tree vs.\ cover tree'' would
reach the same verdict with either library here, but not the same magnitude: on
covtype the KD-tree beats the best cover-tree mode by $3.1\times$ with ours
but by only 7\% with \texttt{kiddo}---precisely the confounding this paper argues against, now shown
outside cover trees. The five-flag KD-tree ablation
(Table~\ref{tab:app_kdtree_ablation}) shows the same structure: only early exit
helps beyond noise, and only at $d{=}784$ ($1.11\times$).

%% ===========================================================================
\section{Conclusion}
\label{sec:conclusion}
%% ===========================================================================

Using \rustknn{} as a controlled experimental platform, we have shown that
cover tree performance depends on interactions between query strategy, $k$,
base parameter, and micro-optimizations in ways not apparent from
cross-codebase comparisons.
The best query mode depends on $k$ and on the dataset---single-tree search at
$k{=}1$ on most of our larger datasets, dual-tree search at $k{\ge}5$ on the
largest---and choosing the wrong mode costs up to $7.5\times$;
base~1.3 is not a safe default for dual-tree search (2.0 or more is optimal in
22 of 24 dataset--$k$ cases);
early-exit distance is the one micro-optimization that matters, and its
benefit grows with dimension ($1.31\times$ at $d{=}54$, $1.49\times$ at
$d{=}784$) while leaving the traversal unchanged;
and all-NN and held-out benchmarks produce different implementation rankings,
so both should be reported. Reporting the main results over 20 independent
allocations with bootstrap intervals is what separated these effects from
measurement noise.

% \paragraph{Limitations.}
Our conclusions are scoped to exact nearest neighbor search under
Euclidean distance, single-threaded, on datasets up to $n{\approx}581$K.
We do not study approximate methods; whether these effects transfer to
HNSW/IVF-style indexes is open, as are non-Euclidean metrics, parallel
scaling, and $n > 10^6$. Timings on the smallest datasets are
noise-limited (Section~\ref{sec:ablation}); cross-codebase timings
carry an instruction-set disparity (AVX-512 vs.\ SSE2;
Appendix~\ref{sec:exp_impl}), though within-\rustknn{} comparisons
are unaffected. Large-dataset times were measured on fully loaded nodes, at
least ${\sim}2\times$ an idle machine's (Appendix~\ref{sec:exp_impl}); we
interpret only ratios measured under the same load.

\section*{Acknowledgment}

We would like to thank Ryan R. Curtin for reviewing early results and helping us sanity check unexpected differences in runtime. Indeed, such issues and discrepancies on what method is fastest, when, and why --- is the kind of confusion we hope to mitigate with this research. 
%% ===========================================================================
% References
%% ===========================================================================

\bibliographystyle{unsrtnat}
\bibliography{references}

%% ===========================================================================
% NeurIPS Paper Checklist
%% ===========================================================================
\clearpage
\section*{NeurIPS Paper Checklist}

\begin{enumerate}

\item {\bf Claims}
    \item[] Question: Do the main claims made in the abstract and introduction accurately reflect the paper's contributions and scope?
    \item[] Answer: \answerYes{}
    \item[] Justification: The abstract and introduction list four numbered contributions (optimization ablation, query mode $\times$ $k$ interaction, base parameter analysis, and all-NN vs.\ held-out divergence) that directly correspond to the experimental sections of the paper.

\item {\bf Limitations}
    \item[] Question: Does the paper discuss the limitations of the work performed by the authors?
    \item[] Answer: \answerYes{}
    \item[] Justification: Yes---the Conclusion contains an explicit ``Limitations'' paragraph. It states that our conclusions are scoped to exact nearest neighbor search under Euclidean distance, single-threaded, on datasets up to $n \approx 581$K; that approximate (HNSW/IVF-style) methods, non-Euclidean metrics, parallel scaling and $n > 10^6$ are not studied; that timings on the smallest datasets are noise-limited (Section~\ref{sec:ablation} draws no micro-optimization conclusions from the five UCI sets with $n \le 351$ or from pima/pendigits); that cross-codebase timings carry an instruction-set disparity; and that absolute large-dataset times were measured on loaded nodes, so only ratios measured under the same load are interpreted.

\item {\bf Theory assumptions and proofs}
    \item[] Question: For each theoretical result, does the paper provide the full set of assumptions and a complete (and correct) proof?
    \item[] Answer: \answerNA{}
    \item[] Justification: This is an empirical systems paper. No new theorems or proofs are presented; the theoretical foundations are from prior work~\cite{izbicki2015faster}.

    \item {\bf Experimental result reproducibility}
    \item[] Question: Does the paper fully disclose all the information needed to reproduce the main experimental results of the paper to the extent that it affects the main claims and/or conclusions of the paper (regardless of whether the code and data are provided or not)?
    \item[] Answer: \answerYes{}
    \item[] Justification: Hardware specifications, software versions, compiler flags, and experimental methodology are described in Section~5.1 and Appendix~B. Implementation details (architecture, packed trees, dual-tree traversal, parallel construction) are in Appendix~C (Sections C.1--C.4). All algorithms are described in sufficient detail to reimplement.

\item {\bf Open access to data and code}
    \item[] Question: Does the paper provide open access to the data and code, with sufficient instructions to faithfully reproduce the main experimental results, as described in supplemental material?
    \item[] Answer: \answerNo{}
    \item[] Justification: The \rustknn{} library---every tree variant, query mode and optimization flag evaluated in the paper, with its tests---is open source under the MIT license at \url{https://github.com/CrowdStrike/RustKNN}. The benchmark harness and cluster job scripts are not released; the measurement protocol (20 independent allocations, estimator, node-load control), hardware, compiler flags and baseline versions are documented in Section~\ref{sec:experiments} and Appendix~\ref{sec:exp_impl}, and all datasets (UCI, MNIST) are public and cited.

\item {\bf Experimental setting/details}
    \item[] Question: Does the paper specify all the training and test details (e.g., data splits, hyperparameters, how they were chosen, type of optimizer) necessary to understand the results?
    \item[] Answer: \answerYes{}
    \item[] Justification: Dataset dimensions and sizes are listed in Table~2, experimental methodology is described in Section~5.1, and hardware/software details including compiler flags are provided in Appendix~B (Section~\ref{sec:exp_impl}).

\item {\bf Experiment statistical significance}
    \item[] Question: Does the paper report error bars suitably and correctly defined or other appropriate information about the statistical significance of the experiments?
    \item[] Answer: \answerYes{}
    \item[] Justification: Every timing table in the main text, and the held-out, waterfall and $k{=}10$ tables in the appendices (Tables~\ref{tab:heldout}, \ref{tab:app_waterfall} and~\ref{tab:allnn_k10}), reports the median over \textbf{20 independent SLURM allocations} with a \textbf{bootstrap 95\% confidence interval} (4{,}000 resamples, seeded by the data so a cell repeated across tables is identical). The base-parameter and per-$k$ ablation tables report the same 20-allocation medians without intervals, and the KD-tree ablation, naive and synthetic appendix tables are medians of several runs within one allocation; each caption says which. The factor of variability captured is between-allocation: each allocation is a separately scheduled job, reduced to one number (the median of its internal timed runs), and the interval is taken over those 20 per-allocation values. Intervals are percentile bootstrap and therefore asymmetric; no normality is assumed.

\item {\bf Experiments compute resources}
    \item[] Question: For each experiment, does the paper provide sufficient information on the computer resources (type of compute workers, memory, time of execution) needed to reproduce the experiments?
    \item[] Answer: \answerYes{}
    \item[] Justification: Experiments were run on a SLURM cluster with Intel Xeon Platinum 8490H processors, one physical core per benchmark job, with node load held at a similar level across campaigns. Hardware and software details are described in the Appendix.
    
\item {\bf Code of ethics}
    \item[] Question: Does the research conducted in the paper conform, in every respect, with the NeurIPS Code of Ethics \url{https://neurips.cc/public/EthicsGuidelines}?
    \item[] Answer: \answerYes{}
    \item[] Justification: The research uses only standard, publicly available machine learning benchmark datasets and raises no ethical concerns.

\item {\bf Broader impacts}
    \item[] Question: Does the paper discuss both potential positive societal impacts and negative societal impacts of the work performed?
    \item[] Answer: \answerNA{}
    \item[] Justification: This work concerns a foundational data structure for nearest neighbor search. It has no direct negative societal impact beyond general-purpose algorithmic research.
    
\item {\bf Safeguards}
    \item[] Question: Does the paper describe safeguards that have been put in place for responsible release of data or models that have a high risk for misuse (e.g., pre-trained language models, image generators, or scraped datasets)?
    \item[] Answer: \answerNA{}
    \item[] Justification: The paper presents a data structure library for nearest neighbor search. No models or datasets with misuse risk are released.

\item {\bf Licenses for existing assets}
    \item[] Question: Are the creators or original owners of assets (e.g., code, data, models), used in the paper, properly credited and are the license and terms of use explicitly mentioned and properly respected?
    \item[] Answer: \answerYes{}
    \item[] Justification: All benchmark datasets (UCI repository, MNIST) are properly cited. Baseline implementations (mlpack~4.7.1, 3-clause BSD license; the C++ reference implementation of Beygelzimer et al., GPL-2.0) and the \texttt{kiddo} Rust crate (MIT or Apache-2.0) are cited, used unmodified apart from JSON output added to the C++ benchmark driver, and are not redistributed with \rustknn{}.

\item {\bf New assets}
    \item[] Question: Are new assets introduced in the paper well documented and is the documentation provided alongside the assets?
    \item[] Answer: \answerYes{}
    \item[] Justification: The new asset is the \rustknn{} library, released at \url{https://github.com/CrowdStrike/RustKNN} under the MIT license, with a README covering installation, usage, the API for each query mode, input limits and feature flags, plus API documentation, tests and runnable examples.

\item {\bf Crowdsourcing and research with human subjects}
    \item[] Question: For crowdsourcing experiments and research with human subjects, does the paper include the full text of instructions given to participants and screenshots, if applicable, as well as details about compensation (if any)?
    \item[] Answer: \answerNA{}
    \item[] Justification: This work does not involve crowdsourcing or research with human subjects.

\item {\bf Institutional review board (IRB) approvals or equivalent for research with human subjects}
    \item[] Question: Does the paper describe potential risks incurred by study participants, whether such risks were disclosed to the subjects, and whether Institutional Review Board (IRB) approvals (or an equivalent approval/review based on the requirements of your country or institution) were obtained?
    \item[] Answer: \answerNA{}
    \item[] Justification: This work does not involve human subjects research.

\item {\bf Declaration of LLM usage}
    \item[] Question: Does the paper describe the usage of LLMs if it is an important, original, or non-standard component of the core methods in this research? Note that if the LLM is used only for writing, editing, or formatting purposes and does \emph{not} impact the core methodology, scientific rigor, or originality of the research, declaration is not required.
    %this research? 
    \item[] Answer: \answerNA{}
    \item[] Justification: LLMs were not used as a component of the core methodology, algorithms, or experimental evaluation in this research.

\end{enumerate}

%% ===========================================================================
\clearpage
\appendix

%% ===========================================================================
\section{Experimental Figures}
\label{sec:app_figures}
%% ===========================================================================

% ---------------------------------------------------------------------------
\FloatBarrier
\subsection{All-Nearest-Neighbor Query Comparison}
\label{sec:app_allnn}

\begin{figure}[h]
  \centering
  \includegraphics[width=\textwidth]{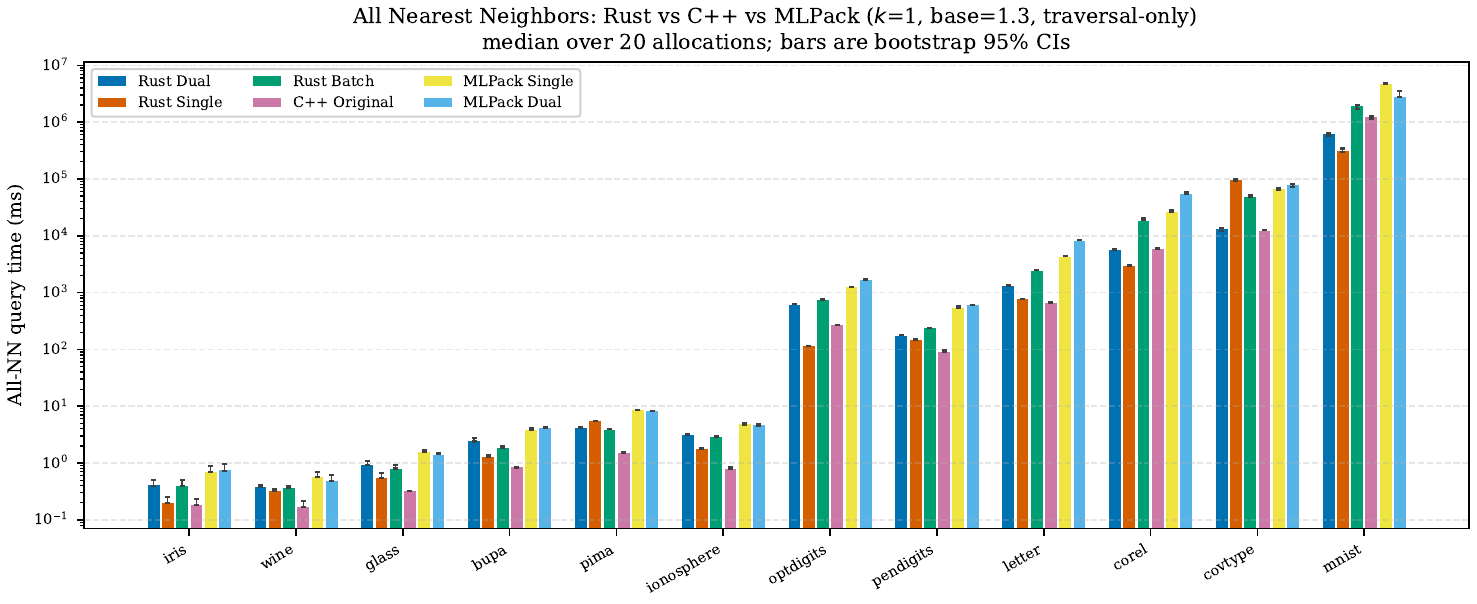}
  \caption{All-nearest-neighbor query time comparison ($k{=}1$, base~1.3).
    Grouped bar chart across all 12~datasets with all six implementations.}
  \label{fig:app_allnn_k1}
\end{figure}

\begin{figure}[h]
  \centering
  \includegraphics[width=\textwidth]{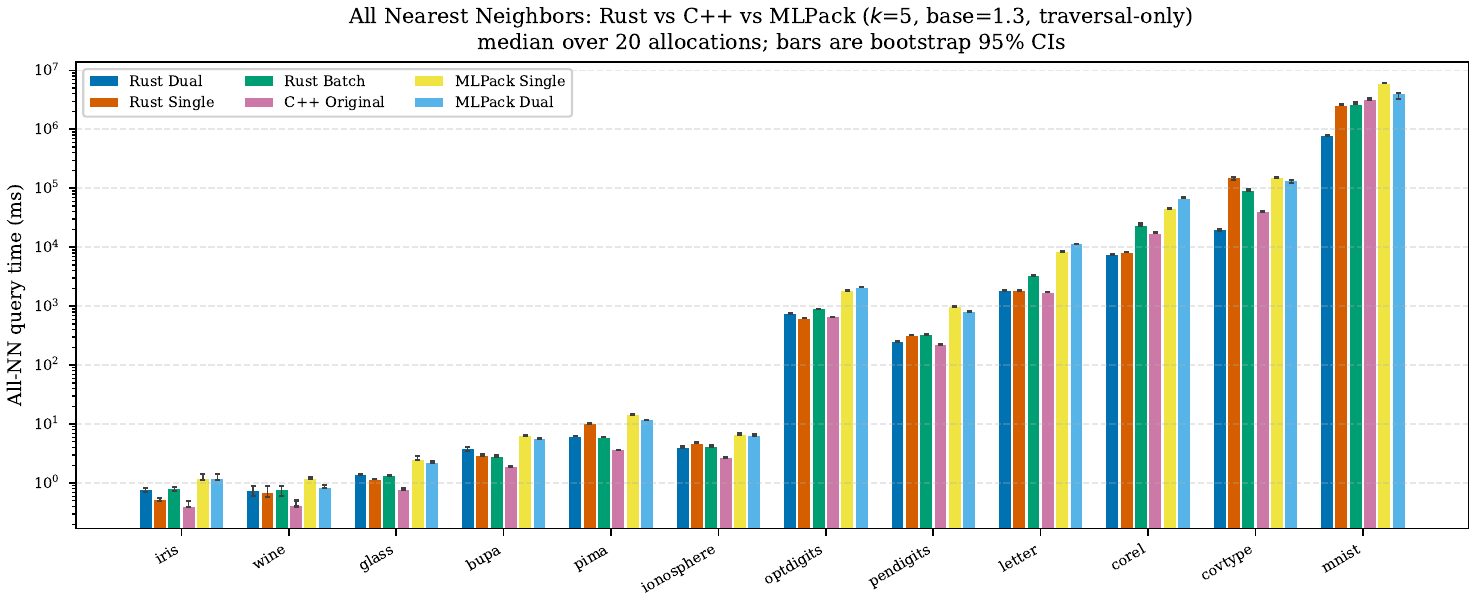}
  \caption{All-nearest-neighbor query time comparison ($k{=}5$, base~1.3).}
  \label{fig:app_allnn_k5}
\end{figure}

\begin{figure}[h]
  \centering
  \includegraphics[width=\textwidth]{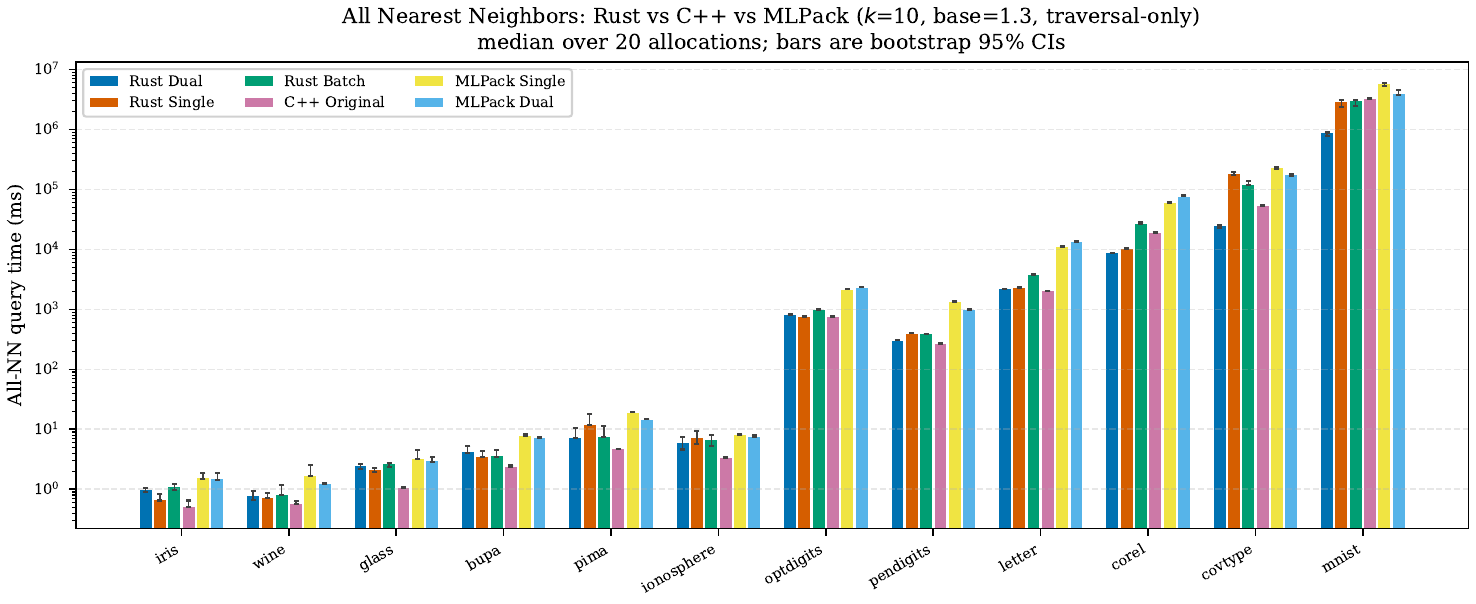}
  \caption{All-nearest-neighbor query time comparison ($k{=}10$, base~1.3).}
  \label{fig:app_allnn_k10}
\end{figure}

Figures~\ref{fig:app_allnn_k1}--\ref{fig:app_allnn_k10} show grouped bar
charts comparing all implementations on the all-nearest-neighbor benchmark
at base~$b{=}1.3$. Bars are medians over 20 independent allocations and
whiskers are bootstrap 95\% confidence intervals, the same estimator used
throughout the paper (Section~\ref{sec:experiments}); on the smallest datasets
the whiskers are wide enough to overlap between implementations, which is the
point made in Section~\ref{sec:ablation}.

% ---------------------------------------------------------------------------
\FloatBarrier
\subsection{Held-Out Query Scaling (Traversal)}
\label{sec:app_heldout_a}

\begin{figure}[h]
  \centering
  \includegraphics[width=\textwidth]{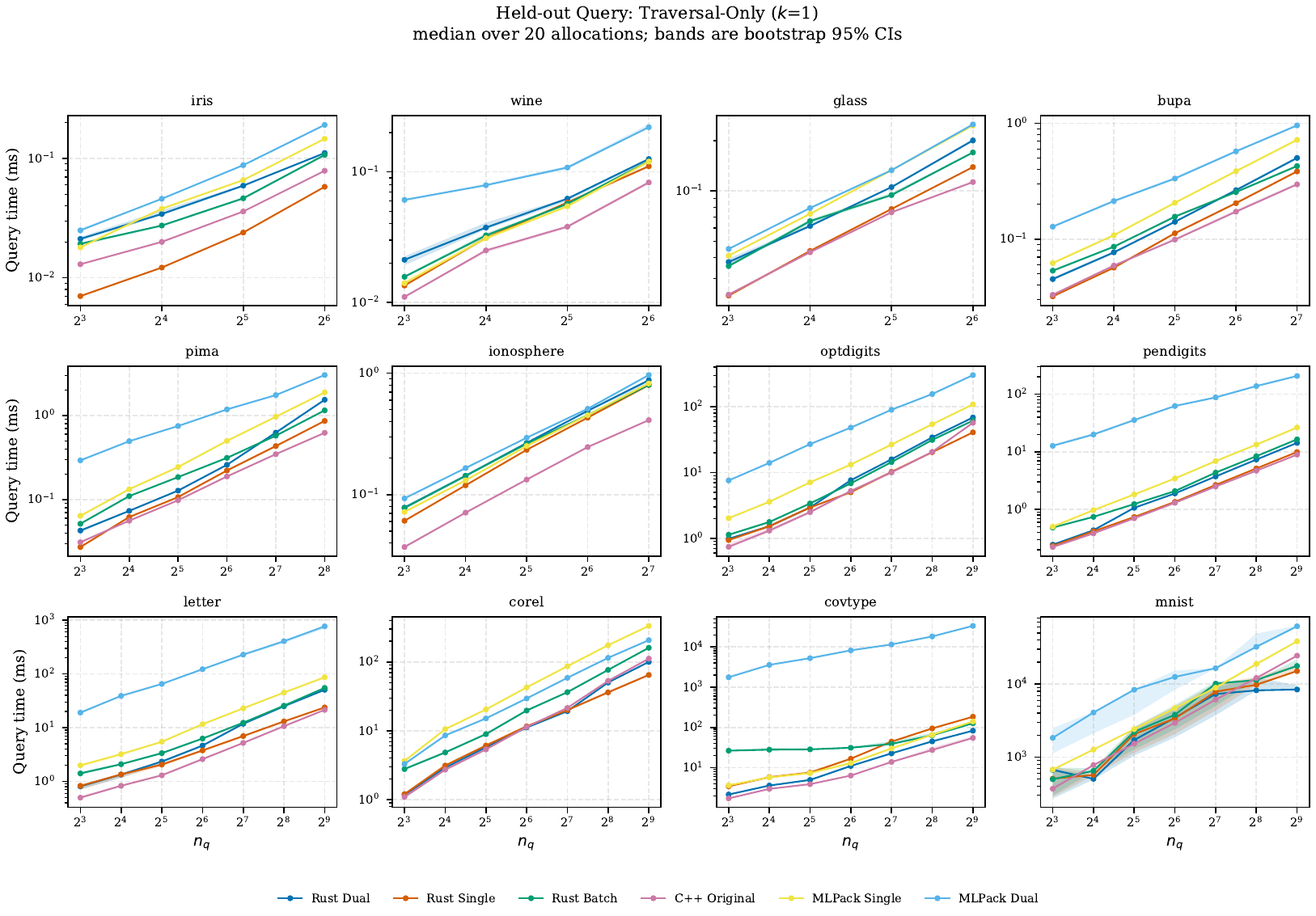}
  \caption{Held-out query scaling, traversal only ($k{=}1$).
    Query time vs.\ number of query points~$n_q$.}
  \label{fig:app_heldout_a_k1}
\end{figure}

\begin{figure}[h]
  \centering
  \includegraphics[width=\textwidth]{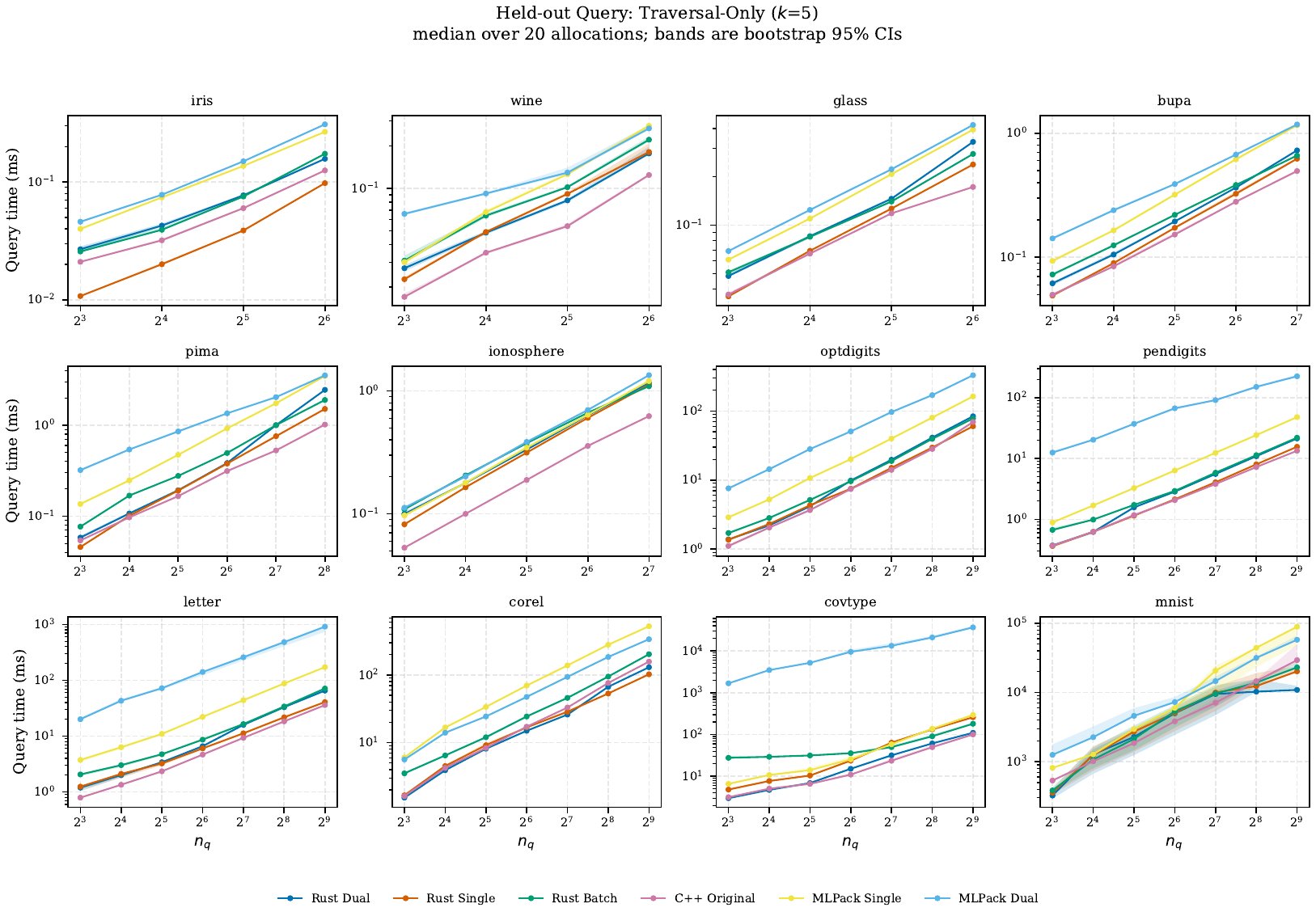}
  \caption{Held-out query scaling, traversal only ($k{=}5$).}
  \label{fig:app_heldout_a_k5}
\end{figure}

\begin{figure}[h]
  \centering
  \includegraphics[width=\textwidth]{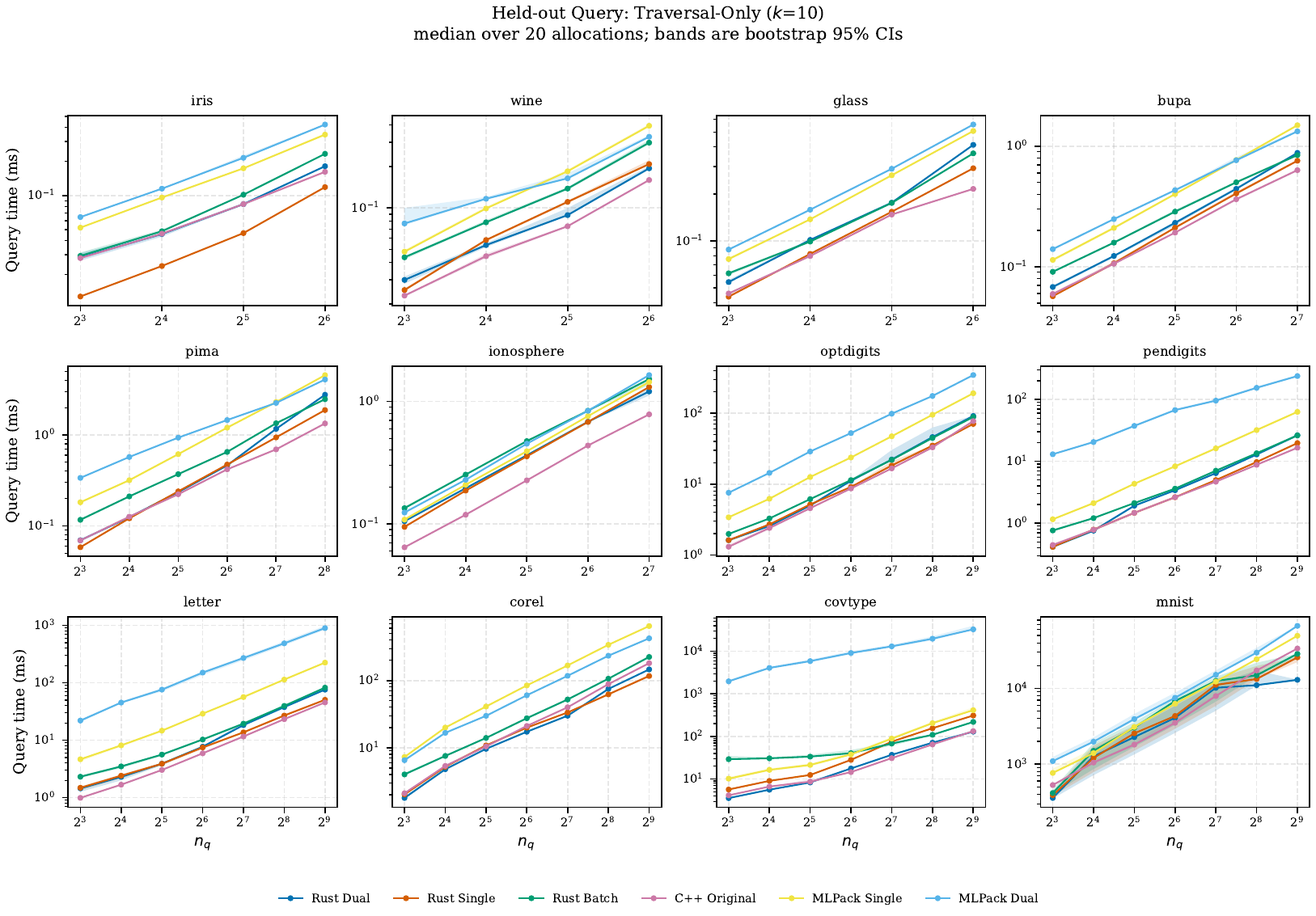}
  \caption{Held-out query scaling, traversal only ($k{=}10$).}
  \label{fig:app_heldout_a_k10}
\end{figure}

\begin{figure}[h]
  \centering
  \includegraphics[width=\textwidth]{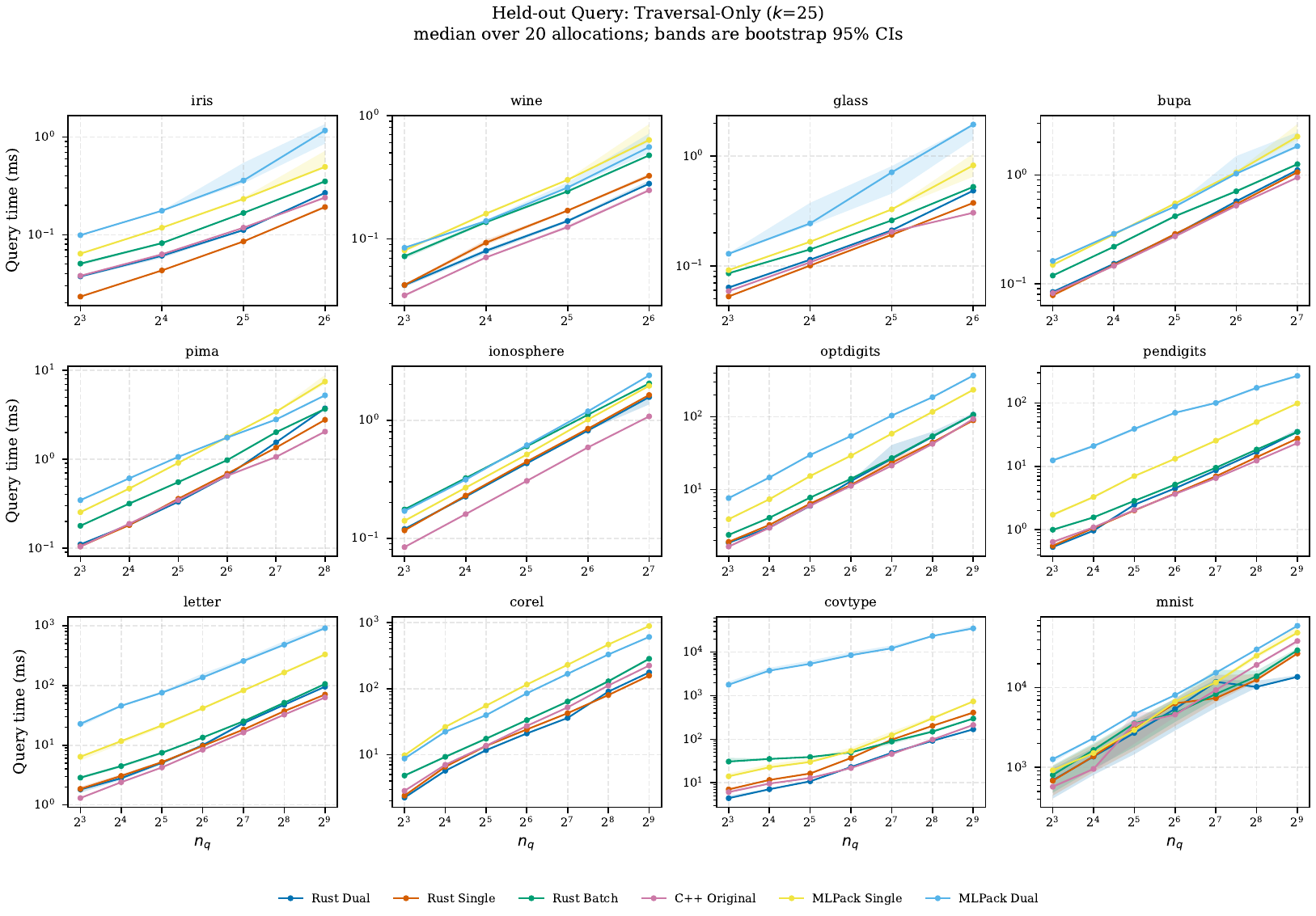}
  \caption{Held-out query scaling, traversal only ($k{=}25$).}
  \label{fig:app_heldout_a_k25}
\end{figure}

\begin{figure}[h]
  \centering
  \includegraphics[width=\textwidth]{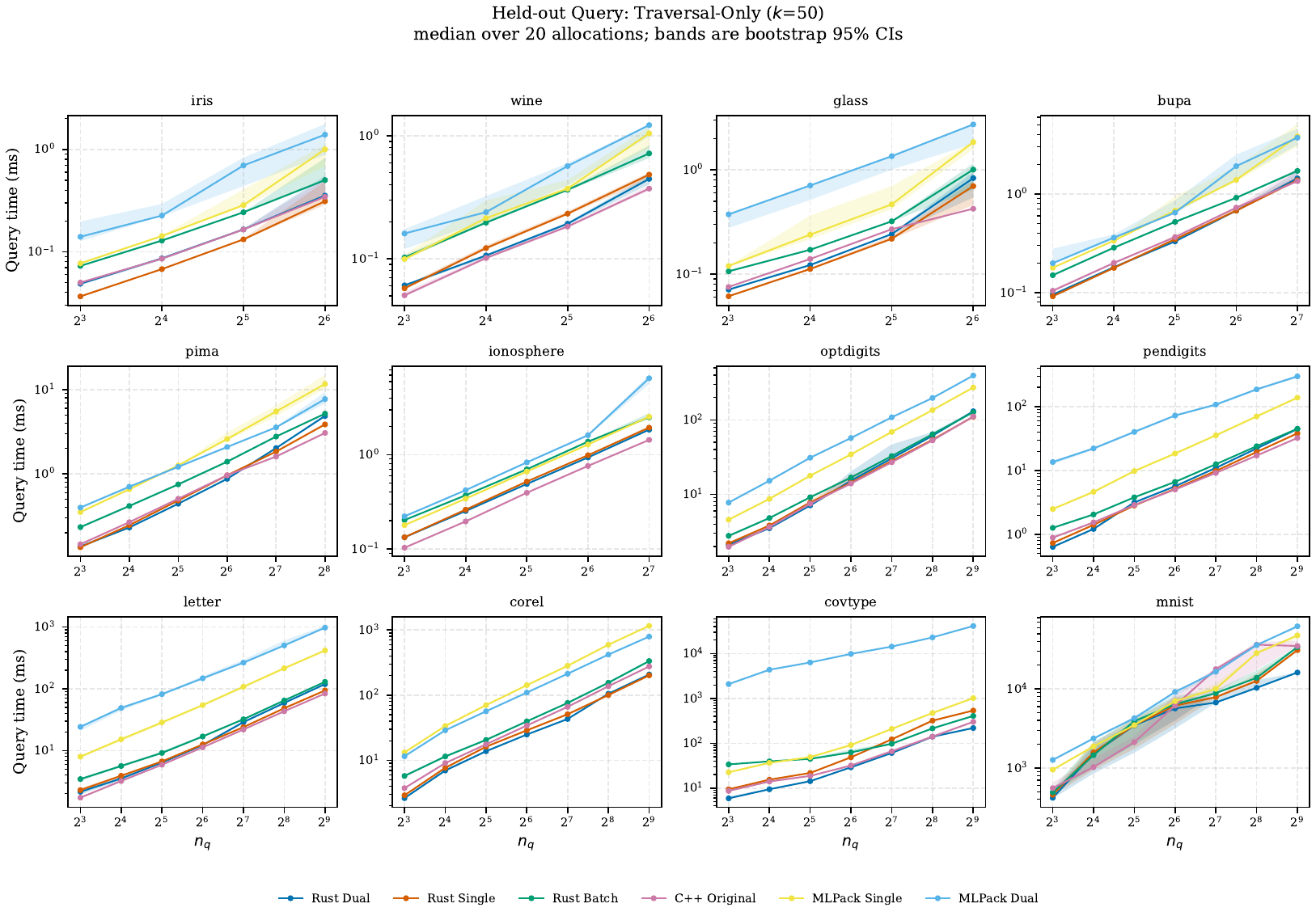}
  \caption{Held-out query scaling, traversal only ($k{=}50$).}
  \label{fig:app_heldout_a_k50}
\end{figure}

Figures~\ref{fig:app_heldout_a_k1}--\ref{fig:app_heldout_a_k50} isolate
traversal-only timing as a function of the number of query points~$n_q$.
Lines are medians over 20 independent allocations; shaded bands are bootstrap
95\% confidence intervals.

\begin{table}[!h]
  \caption{Held-out $k$-NN query time (ms, $k{=}1$) for varying numbers of
    query points~$n_q$. Median [95\% CI] over 20 independent allocations.
    Best per row in \textbf{bold}. pima has $n{=}768$, so its held-out split
    caps at $n_q{=}384$. Values ${\ge}10^3$ are abbreviated K/M.}
  \label{tab:heldout}
  \centering
  \scriptsize
  \setlength{\tabcolsep}{2pt}
  \begin{tabular}{ll rrr rrr}
    \toprule
    & & \multicolumn{3}{c}{\rustknn{} Packed} & C++ & mlpack & mlpack \\
    \cmidrule(lr){3-5}
    Dataset & $n_q$ & Dual & Single & Batch & Ref & Single & Dual \\
    \midrule
    \multirow{3}{*}{pima}
      & 32 & 0.13 [0.11, 0.13] & 0.10 [0.09, 0.10] & 0.17 [0.16, 0.18] & \textbf{0.09 [0.08, 0.09]} & 0.23 [0.22, 0.24] & 0.69 [0.63, 0.70] \\
      & 128 & 0.57 [0.53, 0.57] & 0.39 [0.36, 0.39] & 0.52 [0.49, 0.53] & \textbf{0.31 [0.28, 0.31]} & 0.93 [0.86, 0.94] & 1.64 [1.51, 1.66] \\
      & 384 & 2.17 [2.00, 2.18] & 1.15 [1.05, 1.16] & 1.43 [1.30, 1.43] & \textbf{0.79 [0.73, 0.80]} & 2.72 [2.51, 2.73] & 3.59 [3.31, 3.64] \\
    \midrule
    \multirow{3}{*}{pendigits}
      & 32 & 0.93 [0.79, 1.05] & \textbf{0.59 [0.52, 0.67]} & 1.05 [0.93, 1.16] & 0.63 [0.52, 0.67] & 1.66 [1.41, 1.78] & 31.0 [27.7, 34.1] \\
      & 128 & 3.35 [2.97, 3.55] & 2.32 [2.01, 2.49] & 3.92 [3.45, 4.19] & \textbf{2.22 [1.97, 2.34]} & 6.27 [5.51, 6.79] & 76.1 [69.3, 81.7] \\
      & 512 & 13.2 [13.0, 13.6] & 9.04 [8.86, 9.30] & 15.1 [14.8, 15.5] & \textbf{8.11 [7.26, 8.27]} & 23.6 [21.2, 25.0] & 183 [166, 193] \\
    \midrule
    \multirow{3}{*}{corel}
      & 32 & 5.76 [4.45, 7.28] & 6.10 [4.63, 7.96] & 8.89 [6.89, 11.0] & \textbf{4.70 [3.97, 5.59]} & 19.2 [15.1, 23.9] & 15.2 [11.9, 18.7] \\
      & 128 & 19.3 [15.6, 22.7] & 19.4 [15.6, 23.9] & 35.4 [29.3, 43.0] & \textbf{19.2 [16.0, 21.6]} & 79.4 [62.0, 88.3] & 60.8 [47.5, 66.4] \\
      & 512 & 102 [92.5, 107] & \textbf{65.6 [53.3, 73.4]} & 163 [133, 178] & 104 [83.0, 108] & 300 [231, 331] & 196 [161, 227] \\
    \midrule
    \multirow{3}{*}{covtype}
      & 32 & 4.59 [3.65, 5.81] & 6.90 [5.44, 8.83] & 26.5 [24.4, 33.4] & \textbf{3.83 [3.02, 4.75]} & 7.57 [5.65, 9.70] & 5.84K [4.38, 6.76] \\
      & 128 & 21.8 [16.8, 25.1] & 41.4 [30.8, 48.8] & 37.4 [32.4, 43.6] & \textbf{14.7 [11.1, 16.9]} & 32.9 [23.7, 38.4] & 12.8K [9.54, 14.7] \\
      & 512 & 81.1 [69.6, 90.8] & 176 [127, 203] & 117 [95.3, 136] & \textbf{53.0 [41.6, 60.0]} & 142 [105, 166] & 35.6K [27.4, 41.4] \\
    \bottomrule
  \end{tabular}
\end{table}

% ---------------------------------------------------------------------------
\FloatBarrier
\subsection{Held-Out Query Scaling (End-to-End)}
\label{sec:app_heldout_b}

\begin{figure}[h]
  \centering
  \includegraphics[width=\textwidth]{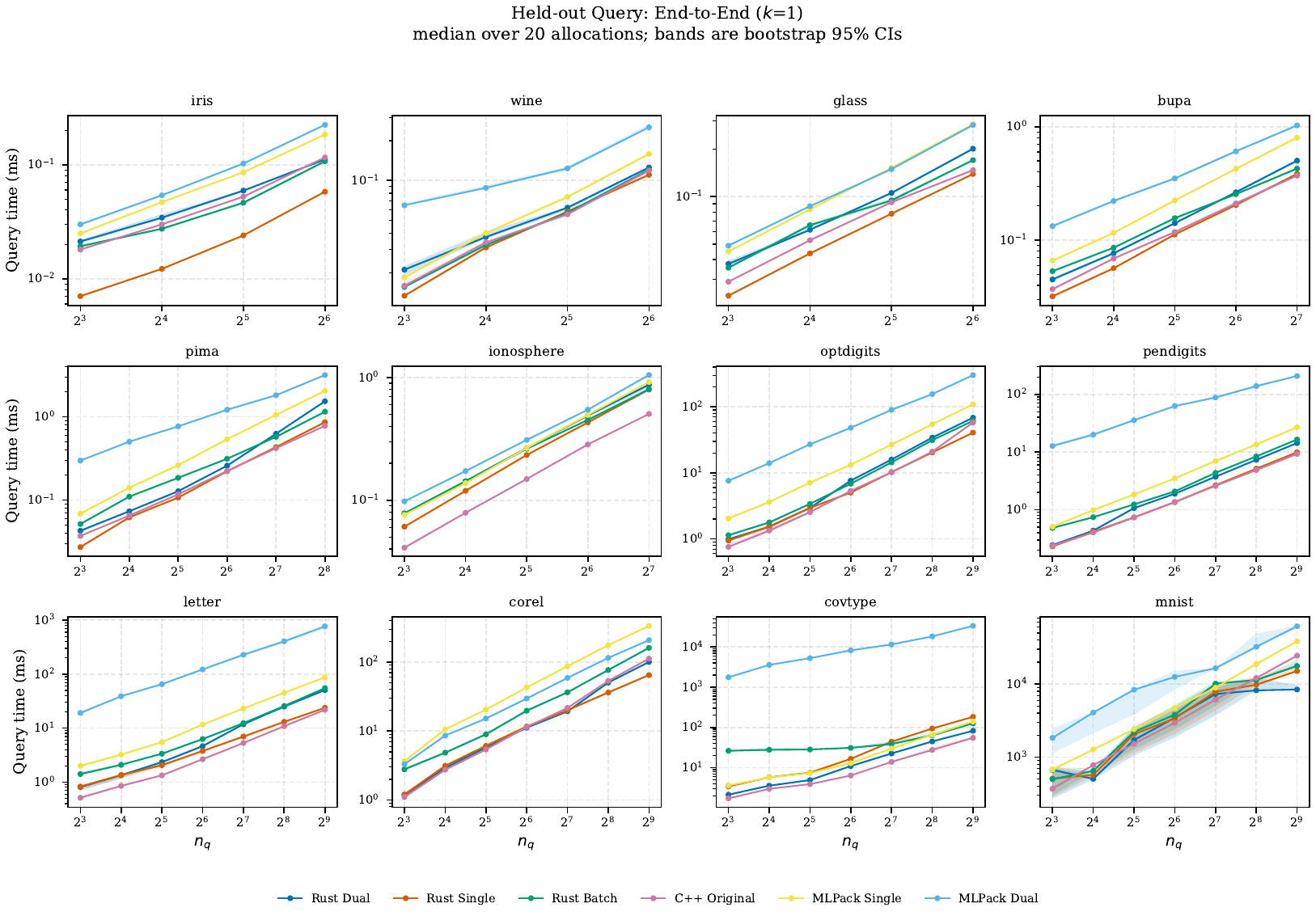}
  \caption{Held-out query scaling, end-to-end ($k{=}1$).
    Query time vs.\ number of query points~$n_q$.}
  \label{fig:app_heldout_b_k1}
\end{figure}

\begin{figure}[h]
  \centering
  \includegraphics[width=\textwidth]{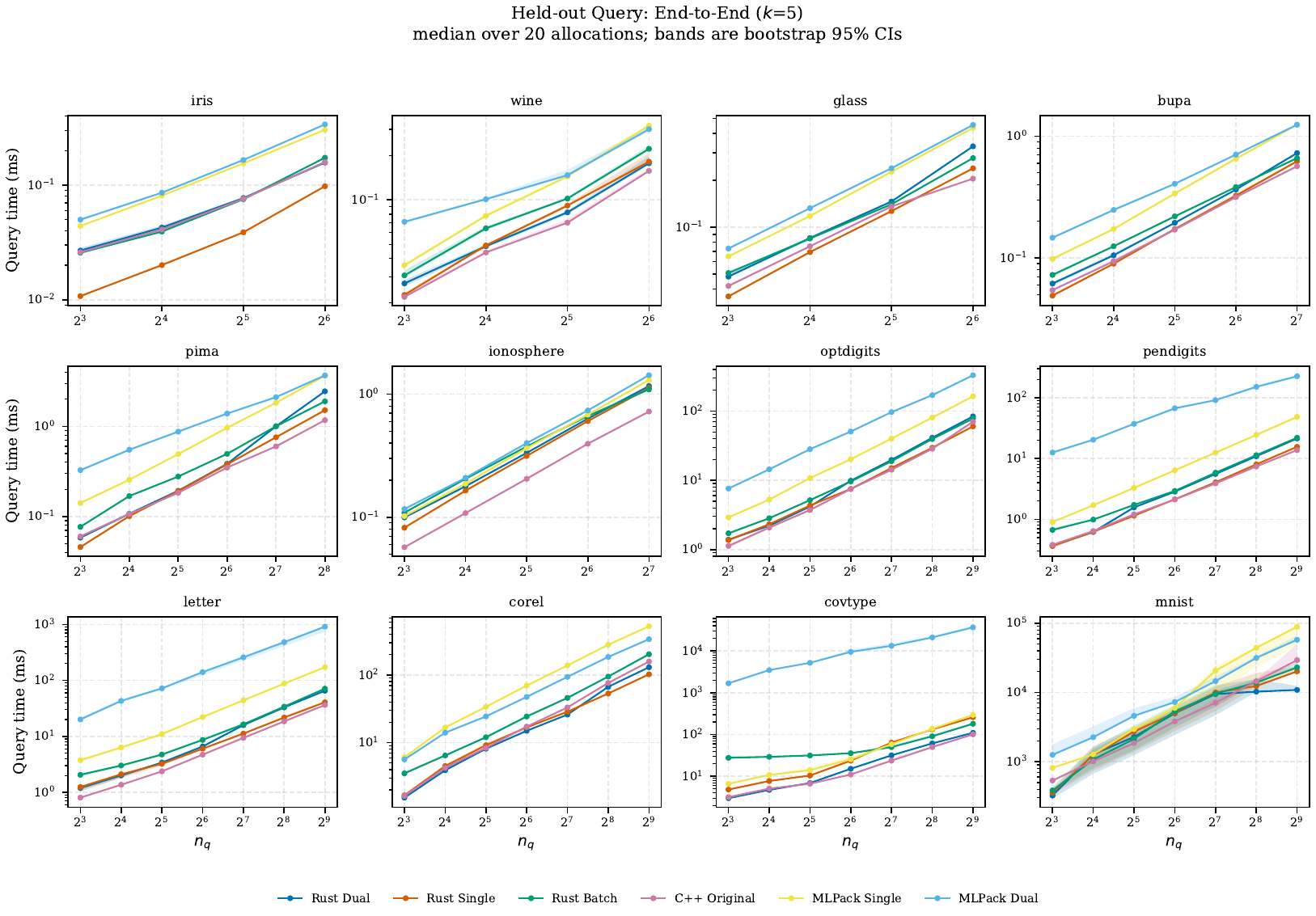}
  \caption{Held-out query scaling, end-to-end ($k{=}5$).}
  \label{fig:app_heldout_b_k5}
\end{figure}

\begin{figure}[h]
  \centering
  \includegraphics[width=\textwidth]{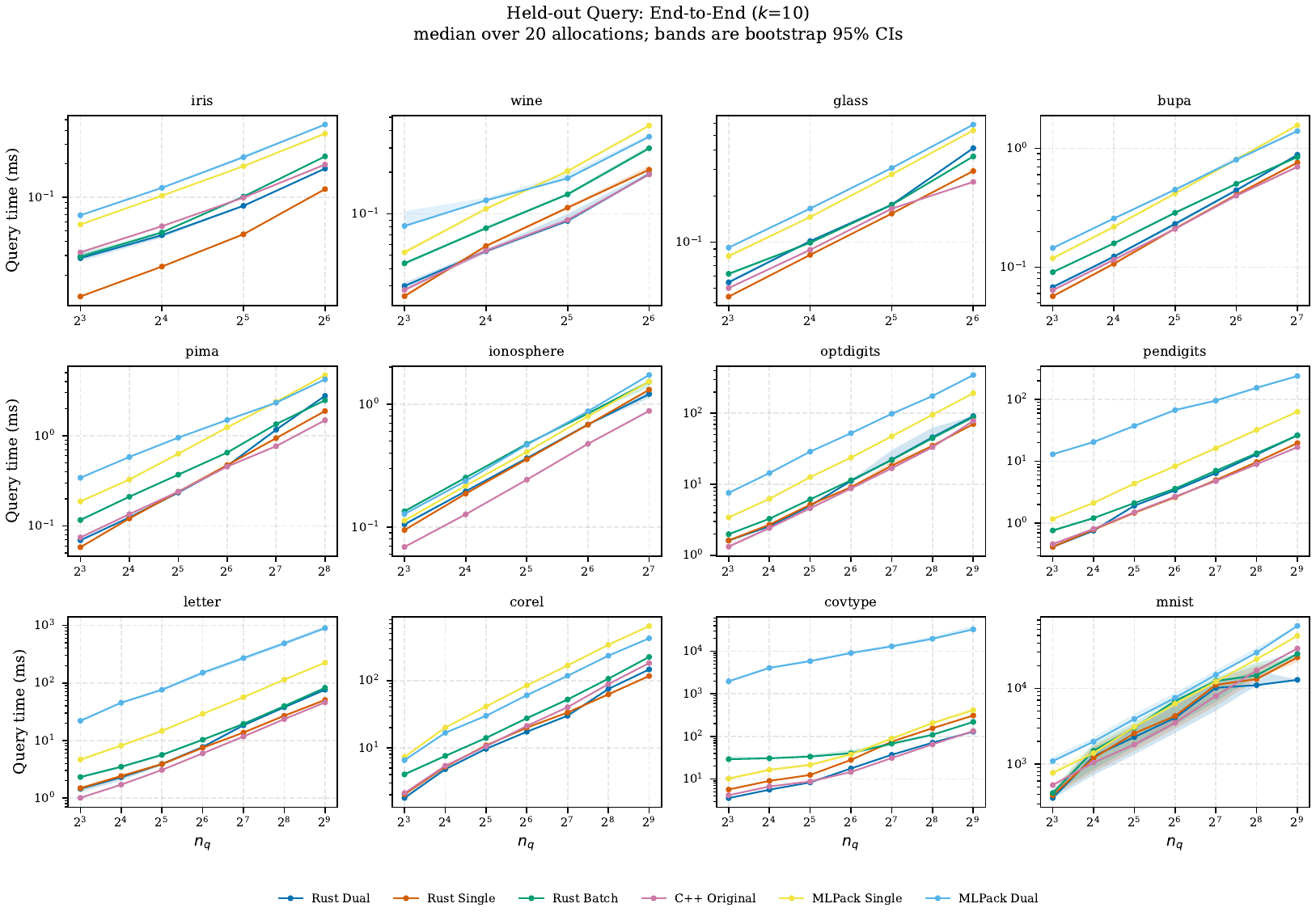}
  \caption{Held-out query scaling, end-to-end ($k{=}10$).}
  \label{fig:app_heldout_b_k10}
\end{figure}

\begin{figure}[h]
  \centering
  \includegraphics[width=\textwidth]{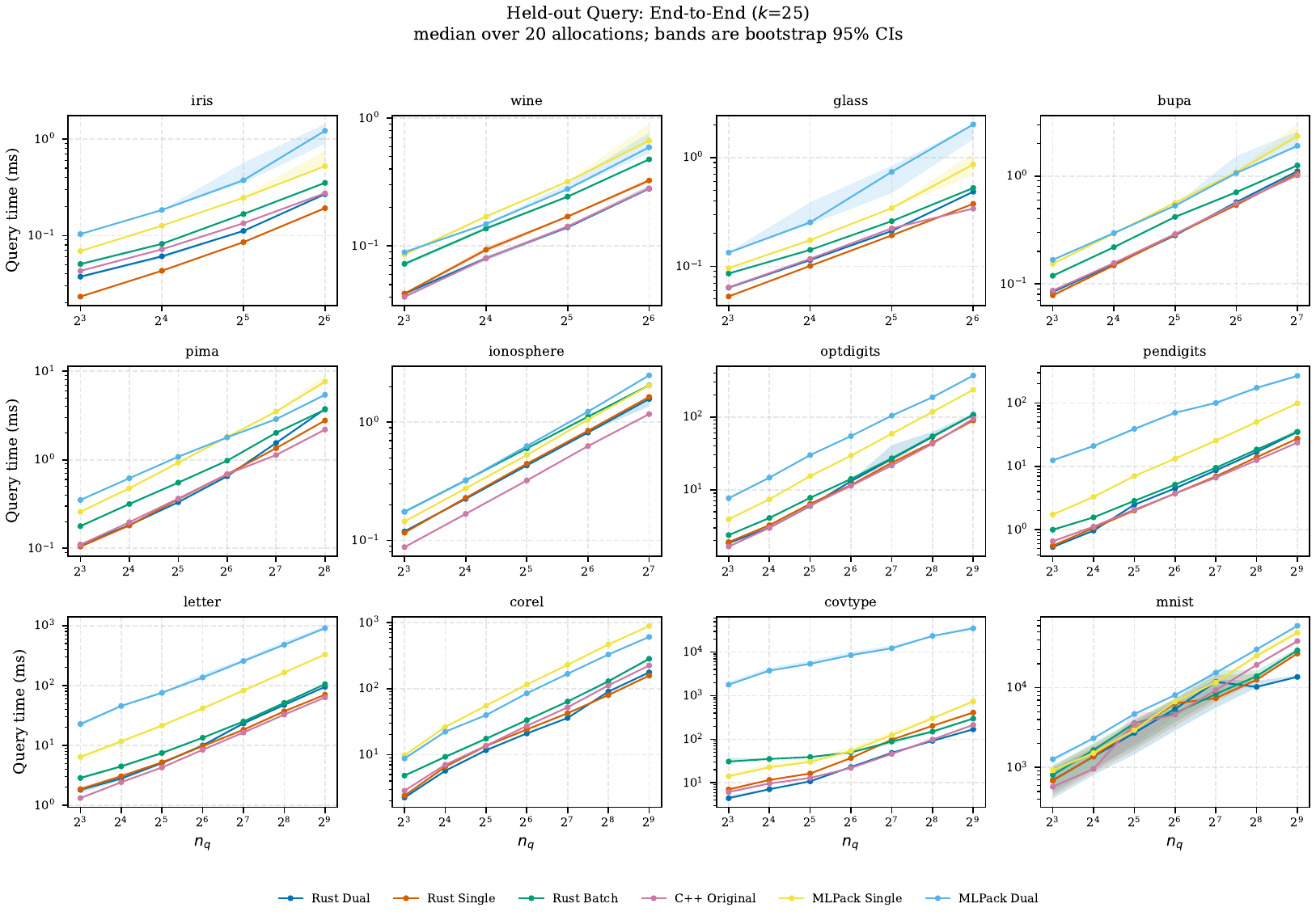}
  \caption{Held-out query scaling, end-to-end ($k{=}25$).}
  \label{fig:app_heldout_b_k25}
\end{figure}

\begin{figure}[h]
  \centering
  \includegraphics[width=\textwidth]{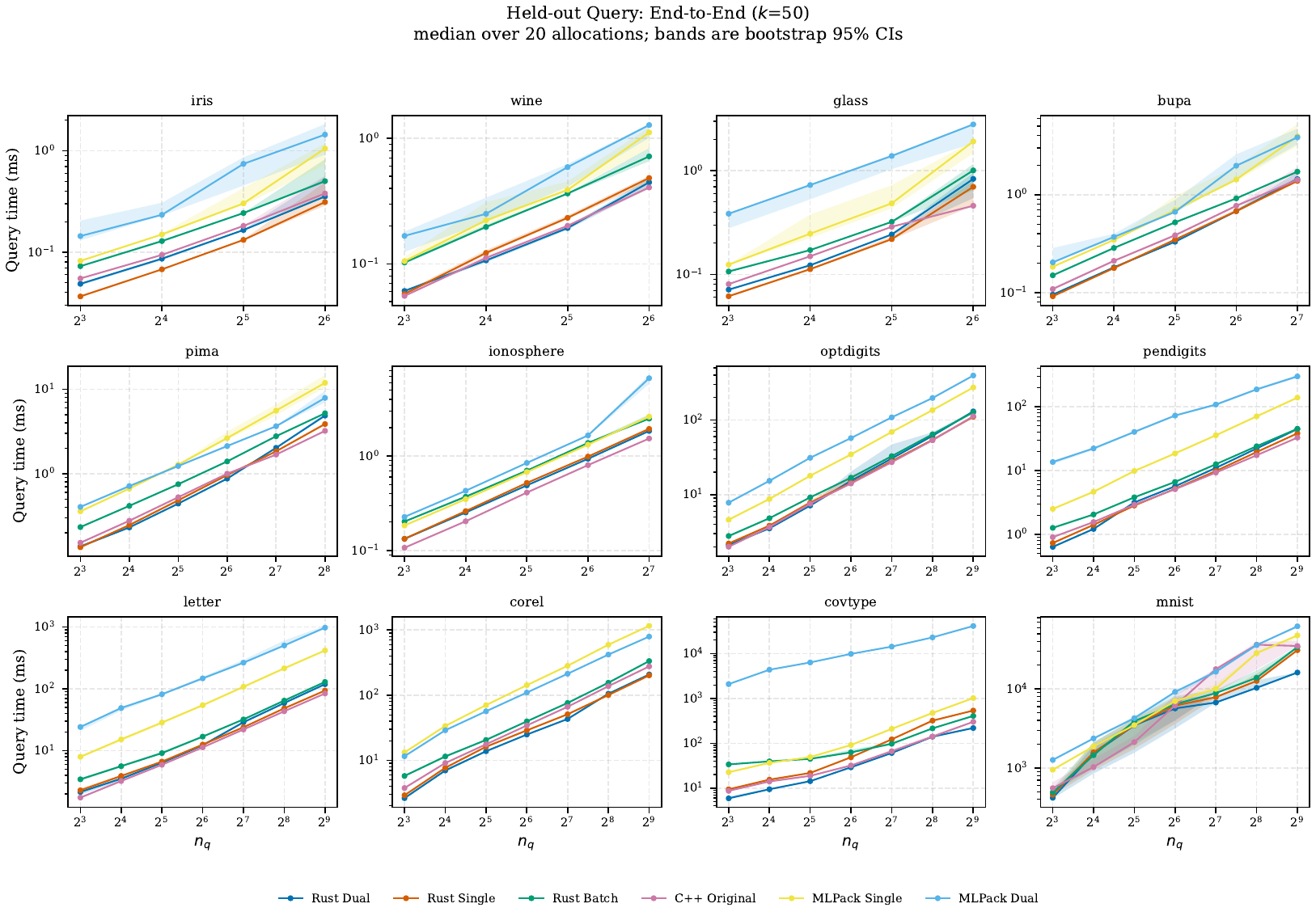}
  \caption{Held-out query scaling, end-to-end ($k{=}50$).}
  \label{fig:app_heldout_b_k50}
\end{figure}

Figures~\ref{fig:app_heldout_b_k1}--\ref{fig:app_heldout_b_k50} show
end-to-end timing (including query-tree construction) as a function of
the number of query points~$n_q$. Lines are medians over 20 independent
allocations; shaded bands are bootstrap 95\% confidence intervals.

% ---------------------------------------------------------------------------
\FloatBarrier
\subsection{Base Parameter Sensitivity}
\label{sec:app_base}

\begin{figure}[h]
  \centering
  \includegraphics[width=\textwidth]{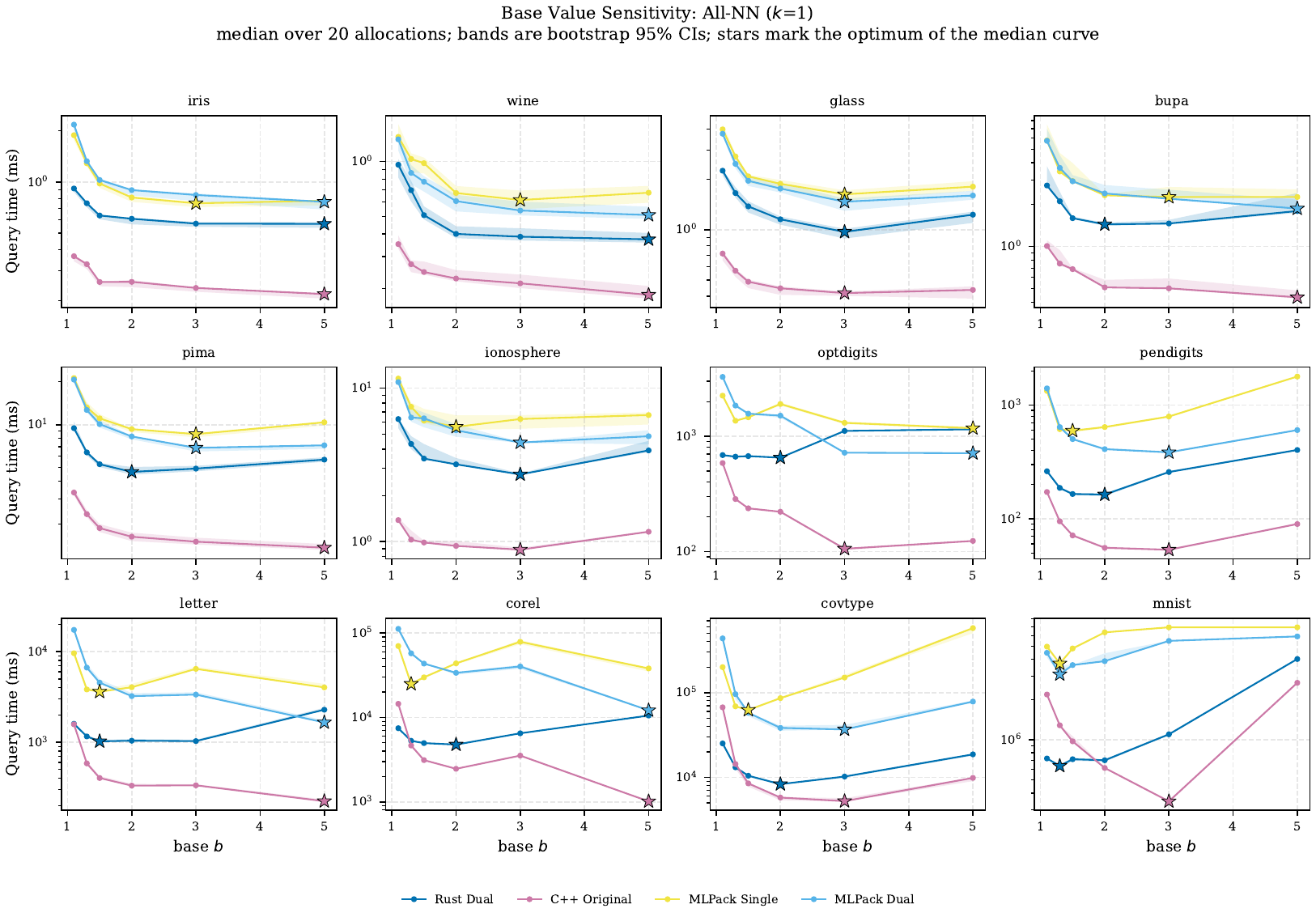}
  \caption{All-nearest-neighbor query time ($k{=}1$) vs.\ base parameter.
    Each panel shows one dataset; implementations are color-coded.
    Medians over 20 independent allocations; shaded bands are bootstrap 95\% CIs;
    stars mark the optimum of each implementation's median curve.
    \rustknn{} curves show traversal time. For \rustknn{} Packed (dual-tree),
    base~1.3 is optimal on only 1 of 12~datasets (mnist) and within 20\% of the
    best base on 5 of 12; elsewhere the optimum is 2.0, 3.0, or (on the two
    smallest sets) 5.0, and 1.5 on letter.}
  \label{fig:base_sensitivity}
\end{figure}

\begin{figure}[h]
  \centering
  \includegraphics[width=\textwidth]{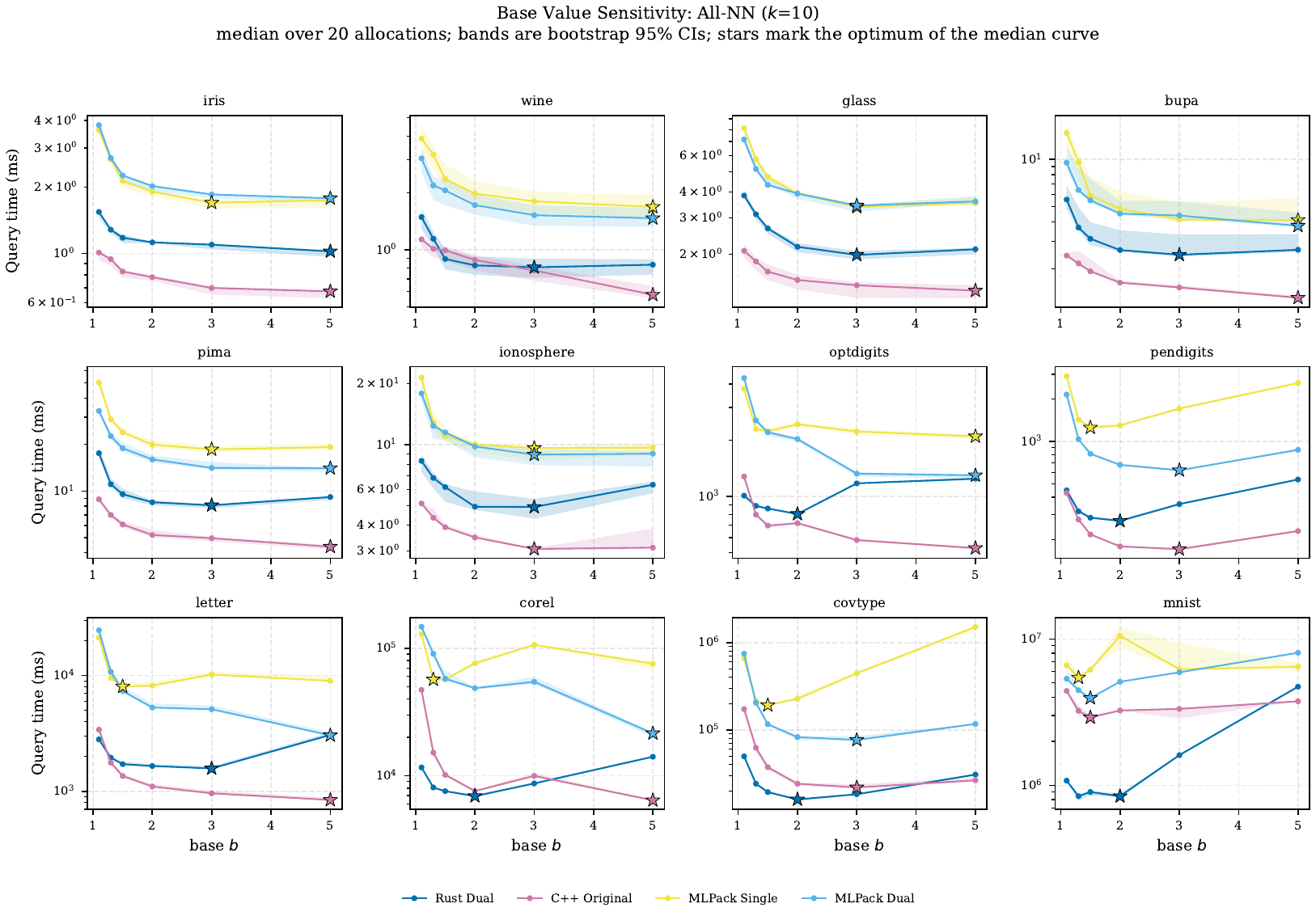}
  \caption{All-nearest-neighbor query time ($k{=}10$) vs.\ base parameter.}
  \label{fig:base_sensitivity_k10}
\end{figure}

\begin{figure}[h]
  \centering
  \includegraphics[width=\textwidth]{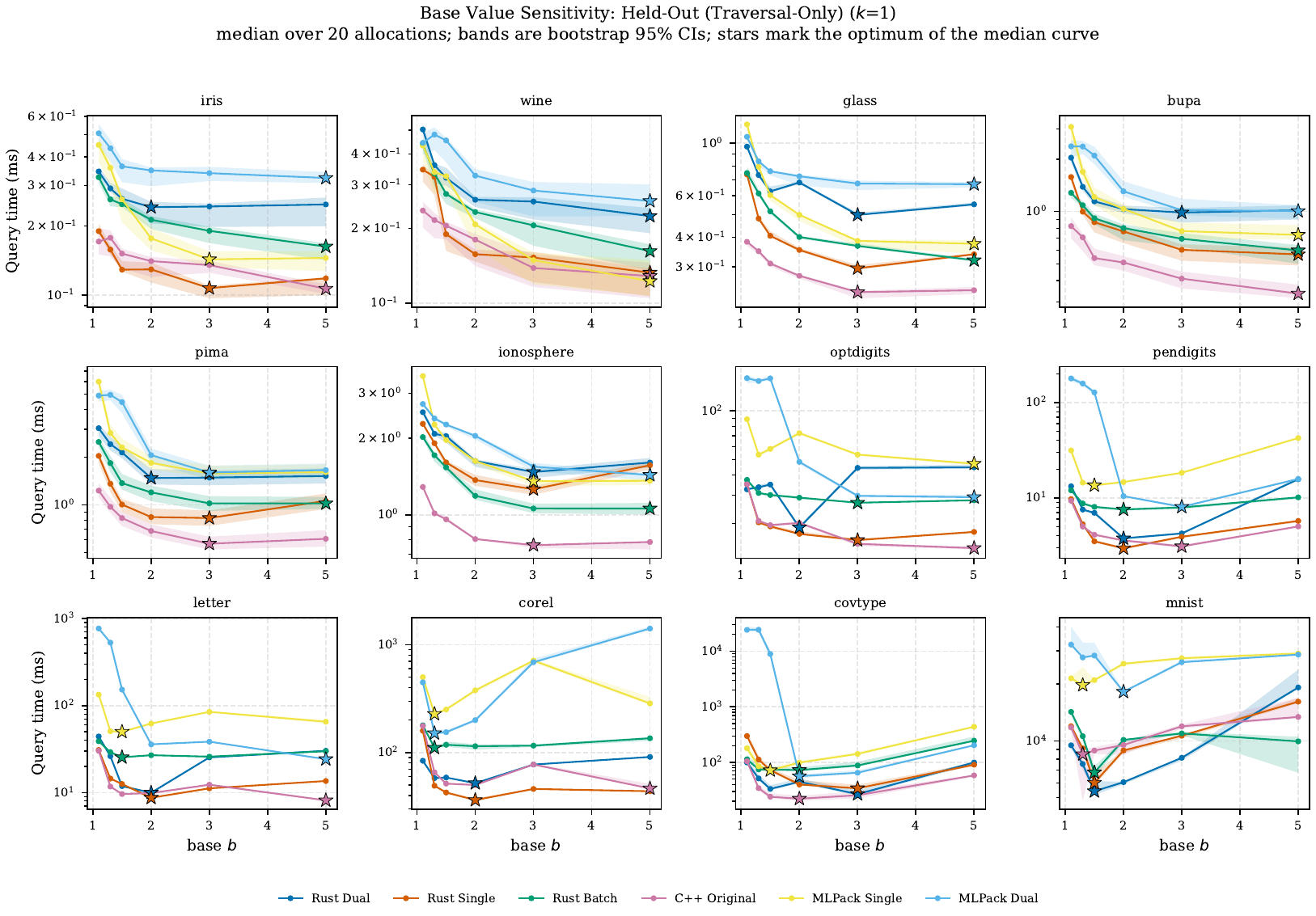}
  \caption{Held-out traversal query time ($k{=}1$) vs.\ base parameter.}
  \label{fig:app_base_heldout_a_k1}
\end{figure}

\begin{figure}[h]
  \centering
  \includegraphics[width=\textwidth]{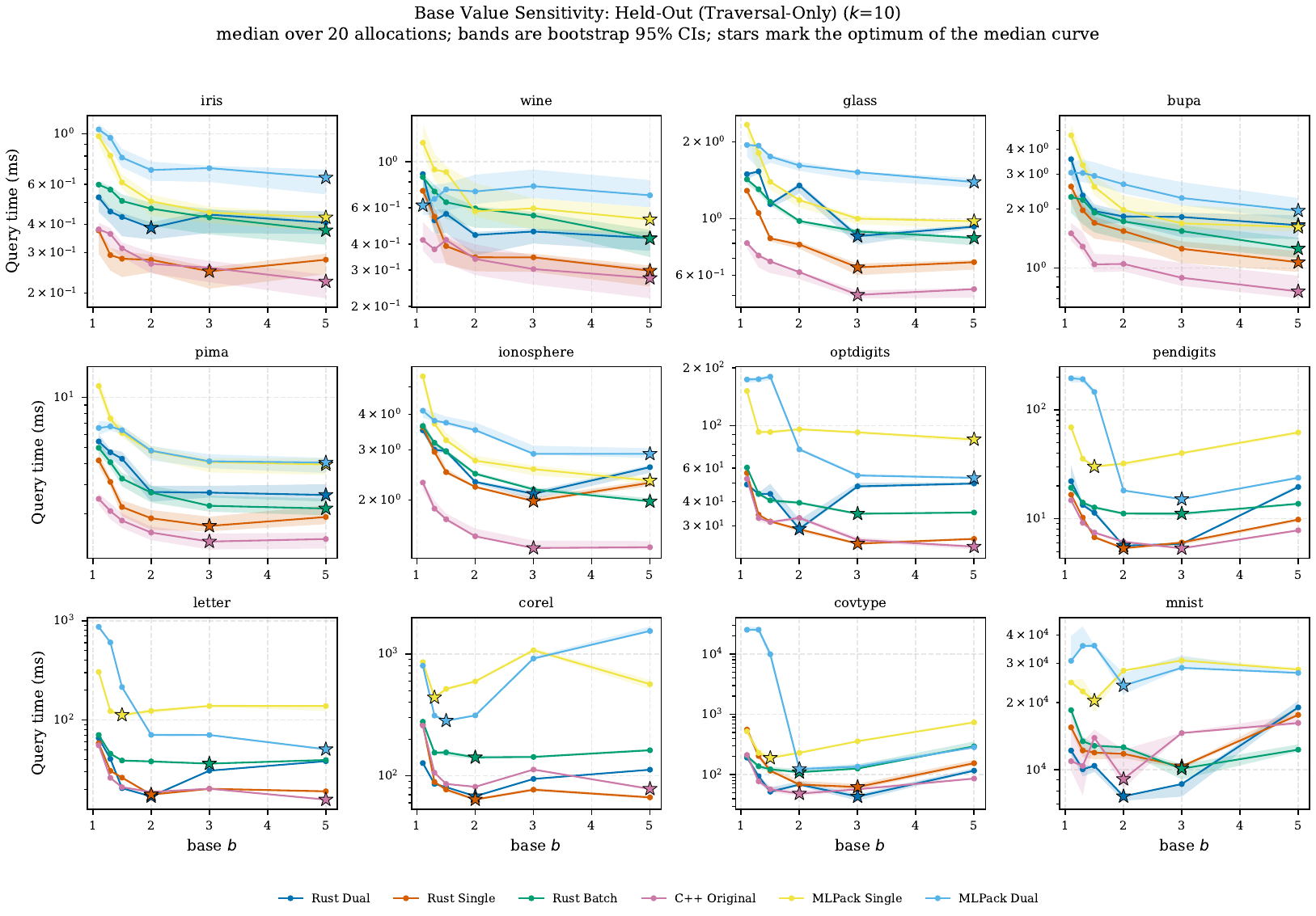}
  \caption{Held-out traversal query time ($k{=}10$) vs.\ base parameter.}
  \label{fig:app_base_heldout_a_k10}
\end{figure}

\begin{figure}[h]
  \centering
  \includegraphics[width=\textwidth]{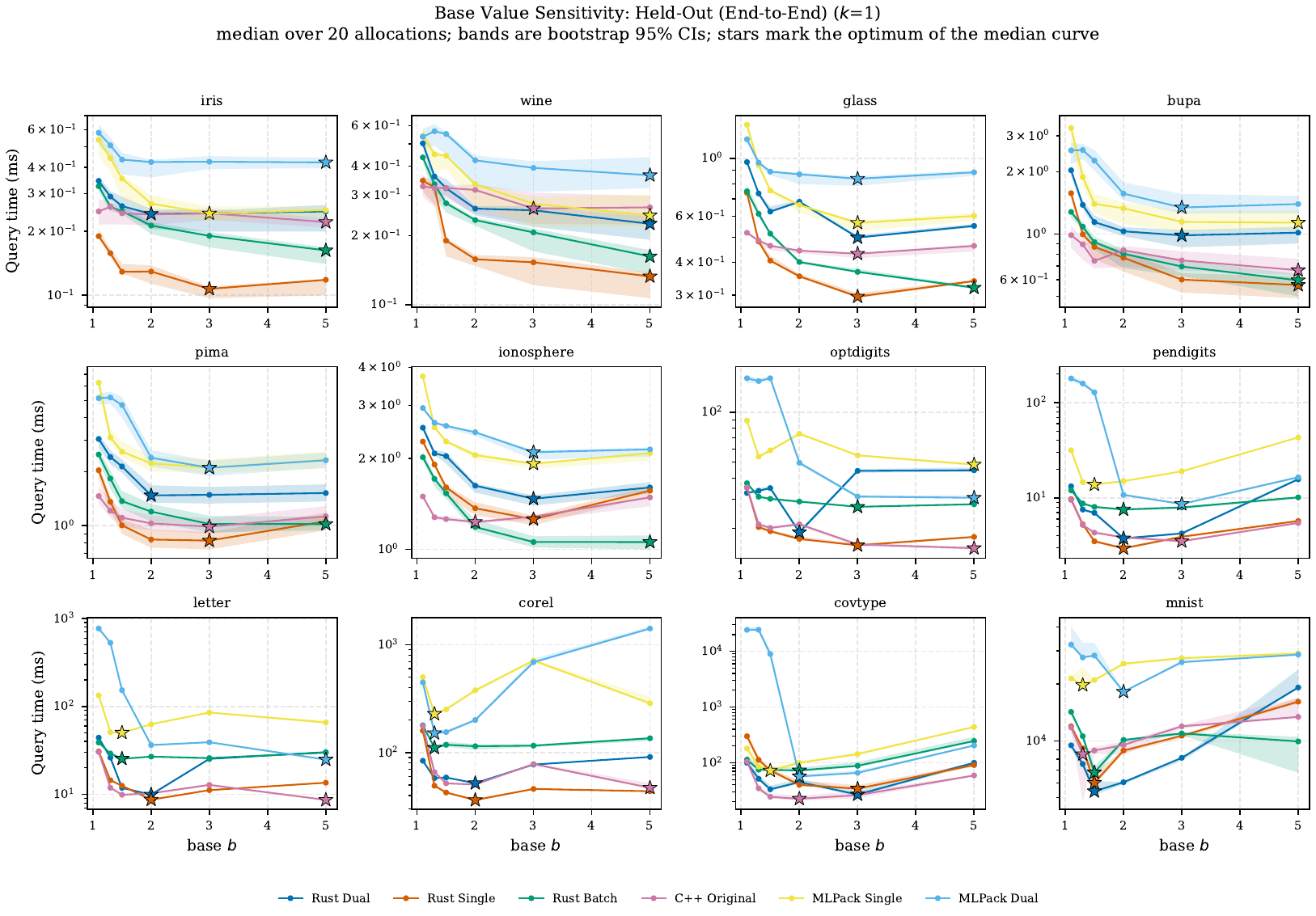}
  \caption{Held-out end-to-end query time ($k{=}1$) vs.\ base parameter.}
  \label{fig:app_base_heldout_b_k1}
\end{figure}

\begin{figure}[h]
  \centering
  \includegraphics[width=\textwidth]{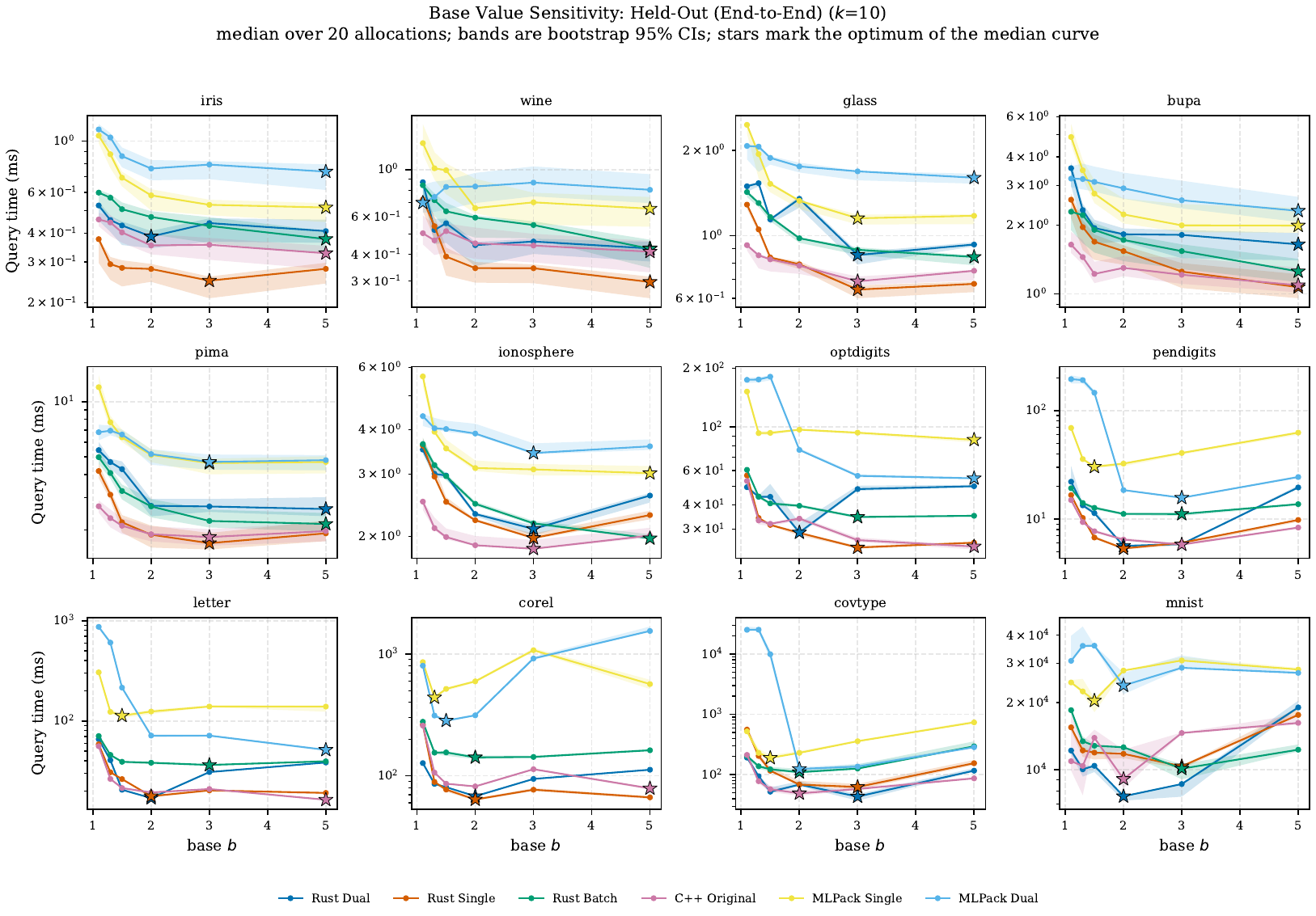}
  \caption{Held-out end-to-end query time ($k{=}10$) vs.\ base parameter.}
  \label{fig:app_base_heldout_b_k10}
\end{figure}

Figures~\ref{fig:base_sensitivity}--\ref{fig:app_base_heldout_b_k10} show
query time as a function of the base parameter
$b \in \{1.1, 1.3, 1.5, 2.0, 3.0, 5.0\}$ for all-nearest-neighbor and held-out
queries. Every panel reports the median over 20 independent allocations with a
bootstrap 95\% confidence interval, the same estimator used throughout the paper.
Note that the held-out sweeps use a query set capped at a fraction of each
dataset, so the smallest sets are evaluated at $n_q < 256$
(iris~75, wine~89, glass~107, bupa~172, ionosphere~175).

\begin{table}[!htb]
  \caption{Optimal base for \rustknn{} Packed dual-tree all-NN, with the cost of
    using the conventional $b{=}1.3$ instead: $t(1.3)/t(b^\star)$, so
    $1.00\times$ means 1.3 is itself optimal. Medians over 20 independent
    allocations from the base-sweep campaign; the absolute times may
    differ from Table~\ref{tab:allnn} due to cross-campaign variance.
    Bold = optimal $b$ differs from~1.3.}
  \label{tab:optimal_base}
  \centering
  \small
  \setlength{\tabcolsep}{3pt}
  \begin{tabular}{l ccc ccc}
    \toprule
    & \multicolumn{3}{c}{$k{=}1$} & \multicolumn{3}{c}{$k{=}10$} \\
    \cmidrule(lr){2-4} \cmidrule(lr){5-7}
    Dataset & Optimal $b$ & Time (ms) & $b{=}1.3$ cost & Optimal $b$ & Time (ms) & $b{=}1.3$ cost \\
    \midrule
    glass       & \textbf{3.0} & 1.04 & 1.67$\times$ & \textbf{3.0} & 2.08 & 1.55$\times$ \\
    ionosphere  & \textbf{3.0} & 2.79 & 1.57$\times$ & \textbf{3.0} & 5.00 & 1.39$\times$ \\
    optdigits   & \textbf{2.0} & 653 & 1.02$\times$ & \textbf{2.0} & 804 & 1.11$\times$ \\
    pendigits   & \textbf{2.0} & 165 & 1.14$\times$ & \textbf{2.0} & 273 & 1.17$\times$ \\
    corel       & \textbf{2.0} & 4{,}773 & 1.11$\times$ & \textbf{2.0} & 6{,}942 & 1.17$\times$ \\
    covtype     & \textbf{2.0} & 8{,}401 & 1.57$\times$ & \textbf{2.0} & 16{,}088 & 1.51$\times$ \\
    mnist       & 1.3 & 638{,}925 & 1.00$\times$ & \textbf{2.0} & 842{,}461 & 1.00$\times$ \\
    \bottomrule
  \end{tabular}
\end{table}

\FloatBarrier
\clearpage
\section{Experiment Details}
\label{sec:exp_impl}

\paragraph{Hardware.}
All experiments run on a multi-node SLURM cluster.
Each compute node contains two Intel Xeon Platinum 8490H processors (Sapphire
Rapids, 60~cores per socket, two hardware threads per core, 112.5\,MiB of L3
per socket, maximum clock 1.90\,GHz, i.e., no turbo boost) with 2\,TB of RAM,
running Ubuntu~22.04 LTS (Linux~5.15).
Every benchmark job is allocated one physical core
(\texttt{-{}-cpus-per-task=1} under core-granularity scheduling) and its own
memory, so jobs never share a core; they do share each socket's L3 cache and
memory bandwidth with the other jobs on the node.

\paragraph{Node load.}
That sharing matters on the large datasets. On an otherwise idle node, $N$
pinned copies of the mnist all-NN, split evenly across the two sockets, slow
down by $1.48\times$ ($N{=}15$), $1.61\times$ ($N{=}30$), and $1.99\times$
($N{=}60$) for the C++ reference, and by $1.43\times$ ($N{=}15$) and
$2.07\times$ ($N{=}60$) for \rustknn{} single-tree search
(each copy pinned to its own physical core): the two degrade at the same
rate, so ratios between implementations measured under the same load are
stable even though absolute times are not. Because the clock is fixed, the
slowdown is cache and memory-bandwidth contention rather than frequency
scaling. We therefore ran every campaign with the nodes held at a similar,
high load (about 110 memory-heavy benchmark jobs per 120-core node), using
filler jobs that run the same mnist workloads while a campaign's last
allocations finished, so that no configuration was measured on an emptying
node. One of the three nodes ran large-dataset allocations $1.1$--$1.5\times$
slower than the other two at the same job count; allocations were spread
across nodes, so this mainly widens the upper confidence bounds on covtype
and mnist rather than moving the medians.
On pima and pendigits, per-allocation timings are bimodal for reasons we
have not isolated (Section~\ref{sec:ablation}).

\paragraph{Software.}
The Rust implementation is compiled with \texttt{rustc}~1.94.1 (stable,
2026-03-25) using \texttt{-{}-release} with LTO enabled and
\texttt{codegen-units=1}.
The reference C++ cover tree of~\citet{beygelzimer2006cover} is compiled
with g++~11.4.0 (\texttt{-O3 -ffast-math -funroll-loops}).
mlpack~4.7.1~\citep{curtin2013mlpack} is compiled with g++~11.4.0
(\texttt{-std=c++17 -O3 -DNDEBUG}).
All implementations use a cover-tree base parameter of~1.3 unless
otherwise noted (sensitivity analysis in Appendix~\ref{sec:app_base}).
Note that the Rust binaries are built on the cluster's login node (Xeon
Platinum 8358, Ice Lake) with \texttt{RUSTFLAGS="-C target-cpu=native"}, i.e.,
AVX-512 code that also runs on the Sapphire Rapids compute nodes, while the
C++ builds use \texttt{g++}'s default instruction set (SSE2); this favors
the Rust implementation on vectorizable distance computations and is
itself a confounder of the kind this paper argues about.
Timing runs use un-instrumented builds; distance counts come from a separate
pass with the \texttt{instrument} build (Section~\ref{sec:experiments}).

%% ===========================================================================
\FloatBarrier
\clearpage
\section{Implementation Details}
\label{sec:app_impl}
%% ===========================================================================

This appendix provides full implementation details for \rustknn{},
supplementing the overview in Section~\ref{sec:implementation}.

\FloatBarrier
\subsection{Architecture and Trait-Based Generics}

The core abstraction enabling arbitrary metric spaces is the \texttt{Distance<T>}
trait:

\begin{lstlisting}[language=Rust, basicstyle=\ttfamily\small]
pub trait Distance<T> {
    fn distance(&self, p: &T, q: &T) -> f64;
    fn distance_with_bound(&self, p: &T, q: &T,
        upper_bound: f64) -> f64 { self.distance(p, q) }
}
\end{lstlisting}

\noindent
Users implement a single method; the library provides optimized built-in
implementations for Euclidean and Manhattan metrics. The optional
\texttt{distance\_with\_bound} method enables early-exit distance
computation: the built-in Euclidean implementation uses four independent
accumulators with 8-element unrolled loops for instruction-level parallelism,
checking partial squared sums every 16--64 dimensions. If the partial sum
exceeds \texttt{upper\_bound}$^2$, the function returns a sentinel value
without computing the final square root.

The generic node structure \texttt{Node<T: Clone>} stores the data point,
level, cached \texttt{maxdist}, cached parent distance \texttt{d\_parent},
children as \texttt{Vec<Box<Node<T>{>}{>}>}, and an \texttt{is\_duplicate} flag
for merge-created duplicates. Fields are ordered for cache locality: hot query
fields (\texttt{point}, \texttt{maxdist}, \texttt{d\_parent}) appear first.

The library is organized into three variant modules
(\texttt{simplified/}, \texttt{nearest\_ancestor/}, \texttt{packed/}) sharing
common algorithms in a \texttt{core/} directory. A zero-cost \texttt{CoverTree}
enum wrapper delegates via pattern matching.

\FloatBarrier
\subsection{Packed Trees and Cache Optimization}

After construction, the pointer-based tree is converted to a
\texttt{PackedCoverTree} via a single depth-first traversal. All nodes are
stored in a contiguous \texttt{Vec<PackedNode<T>{>}}, with child relationships
in a separate shared \texttt{Vec<usize>} array. Each \texttt{PackedNode}
carries \texttt{u32} fields (\texttt{child\_index\_start},
\texttt{child\_index\_count}) as indices into the shared array.

This yields compact nodes: \texttt{PackedNode<f64>} occupies approximately
40~bytes, with hot query fields spanning $\le$32 bytes---fitting within a
single 64-byte cache line. Using \texttt{u32} saves 8 bytes per node versus
\texttt{usize}. Approximately 1{,}000 nodes fit in a 32\,KB L1 cache.

During packing, children are sorted by \texttt{d\_parent} (ascending) so
closer children appear first in memory, tightening the pruning bound sooner.

\FloatBarrier
\subsection{Dual-Tree Traversal}

The dual-tree traversal requires several adaptations for cover trees without
self-children. When expanding one side's children while holding the other
constant, the held-constant node's \emph{point} must still score against the
expanded children. Two helper routines (\texttt{point\_vs\_subtree},
\texttt{subtree\_vs\_point}) simulate the missing self-child.

When both nodes have children, the traversal proceeds in three phases:
(1)~query point $\times$ reference children,
(2)~query children $\times$ reference point, and
(3)~child--child pairs using triangle inequality lower bounds from
phases~1 and~2 as free pre-filters:
$\mathrm{lb} = \max(|d(q_p, r_{c_j}) - d_{\mathrm{parent}}(q_{c_i})|,\;
|d(q_{c_i}, r_p) - d_{\mathrm{parent}}(r_{c_j})|)$.
Pairs whose lower bound minus combined \texttt{maxdist} exceeds the pruning
threshold are skipped without additional distance computation.

Per-query-point bounds are cached in a flat \texttt{Vec} indexed by node
index. Two bound modes are supported: full recursive B1/B2 bounds with
epoch-based cache invalidation (the default), and a Beygelzimer-style bound.
A cheaper per-point $k$th-distance bound was removed: it drops the per-subtree
margin and can prune true neighbors.

\FloatBarrier
\subsection{Parallel Construction}

Parallel construction follows the partition-build-merge strategy
of~\citet{izbicki2015faster}. The dataset is partitioned into equal chunks,
each thread builds an independent cover tree, and trees are merged via binary
reduction in $O(\log p)$ rounds. The merge algorithm aligns tree levels,
partitions children into covered and uncovered sets, recursively merges
covered children, and reinserts uncovered points. Exact \texttt{maxdist}
recomputation after merge restores optimal pruning bounds.

%% ===========================================================================
\FloatBarrier
\clearpage
\section{Full Ablation Tables}
\label{sec:app_ablation}
%% ===========================================================================

Table~\ref{tab:app_waterfall} shows the cumulative waterfall: starting from
``all off'' (all five optimizations disabled), each row adds one optimization.
Values are speedup over all-off ($\times$); the final ``Baseline'' row gives
the combined speedup of all optimizations.
This table and Table~\ref{tab:ablation} report the same endpoint quantity,
$t(\text{all off})/t(\text{baseline})$, and are computed from a
\emph{single pooled 20-allocation sample}, so the ``Baseline'' row below is
identical to Table~\ref{tab:ablation}'s ``all off'' row by construction rather
than merely up to noise.

\begin{table}[h]
  \caption{Cumulative waterfall: speedup ($\times$ all-off) as optimizations
    are added one at a time ($k{=}1$, all-NN, base~1.3).
    Median [95\% CI] over 20 independent allocations.
    Optimizations are added in order of increasing sophistication.
    The ``Baseline'' row equals the ``all off'' row of
    Table~\ref{tab:ablation} by construction (same pooled sample).}
  \label{tab:app_waterfall}
  \centering
  \footnotesize
  \setlength{\tabcolsep}{3pt}
  \begin{tabular}{l ccccc}
    \toprule
    Configuration & pima & pendigits & corel & covtype & mnist \\
    \midrule
    All off        & 1.00 [0.83, 1.21] & 1.00 [0.93, 1.08] & 1.00 [0.97, 1.03] & 1.00 [0.98, 1.02] & 1.00 [0.88, 1.13] \\
    +Triangle      & 0.88 [0.77, 1.14] & 1.00 [0.92, 1.03] & 1.02 [0.99, 1.03] & 1.03 [1.01, 1.04] & 1.12 [0.98, 1.13] \\
    +$k{=}1$ fast  & 1.14 [0.97, 1.52] & 0.99 [0.92, 1.03] & 1.01 [0.96, 1.02] & 1.02 [1.01, 1.03] & 1.12 [0.98, 1.16] \\
    +SmallVec      & 0.91 [0.79, 1.26] & 1.00 [0.93, 1.04] & 1.01 [0.99, 1.03] & 1.06 [1.02, 1.09] & 1.12 [0.98, 1.14] \\
    +Halfsort      & 1.24 [0.87, 1.59] & 0.96 [0.78, 1.23] & 1.08 [1.05, 1.09] & 1.02 [0.97, 1.05] & 1.06 [0.86, 1.14] \\
    +Early-exit    & 1.11 [0.99, 1.31] & 0.78 [0.72, 0.83] & 1.06 [1.03, 1.07] & 1.33 [1.31, 1.34] & 1.58 [1.39, 1.60] \\
    Baseline       & 1.11 [0.99, 1.31] & 0.78 [0.72, 0.83] & 1.06 [1.03, 1.07] & 1.33 [1.31, 1.34] & 1.58 [1.39, 1.60] \\
    \bottomrule
  \end{tabular}
\end{table}

On covtype ($d{=}54$) and mnist ($d{=}784$), nearly the entire combined
effect comes from the final step: adding early-exit raises the speedup from
$1.02\times$ to $1.33\times$ on covtype and from $1.06\times$ to $1.58\times$
on mnist, while the four traversal-level optimizations before it contribute at
most about 12\% (on mnist, within intervals that include~1.0).
On corel ($d{=}32$) the combined effect is only $1.06\times$, and adding
early-exit does not increase it.
The composition is largely additive on the large datasets: early-exit
targets the cost of each distance computation while the other four target
traversal overhead, so they operate on independent bottlenecks.
On pima and pendigits the per-allocation timings are bimodal
(Section~\ref{sec:ablation}), so no ordering of the intermediate rows on
those two datasets is meaningful; we report them for completeness only.

Table~\ref{tab:app_ablation_perk} shows the full per-$k$ individual
ablation results for the three larger datasets, where the intervals in
Table~\ref{tab:ablation} are tight enough for the numbers to carry meaning.
Its $k{=}1$ column is the same quantity as Table~\ref{tab:ablation}'s
corresponding entries and is computed from the same 20-allocation source, so
the two agree by construction.

\begin{table}[h]
  \caption{Per-$k$ ablation slowdown factors ($\times$ baseline, all-NN,
    base~1.3) for corel, covtype, and mnist. Values ${>}1$ indicate the
    optimization helps. Median over 20 independent allocations at every~$k$;
    the $k{=}1$ column matches Table~\ref{tab:ablation} by construction.
    Per-cell confidence intervals are omitted for width and are reported in
    the supplementary analysis output.}
  \label{tab:app_ablation_perk}
  \centering
  \small
  \begin{tabular}{l cccc cccc cccc}
    \toprule
    & \multicolumn{4}{c}{corel ($d{=}32$)}
    & \multicolumn{4}{c}{covtype ($d{=}54$)}
    & \multicolumn{4}{c}{mnist ($d{=}784$)} \\
    \cmidrule(lr){2-5} \cmidrule(lr){6-9} \cmidrule(lr){10-13}
    Disabled & $k{=}1$ & $5$ & $10$ & $50$
             & $k{=}1$ & $5$ & $10$ & $50$
             & $k{=}1$ & $5$ & $10$ & $50$ \\
    \midrule
    Early-exit     & 0.98 & 0.99 & 0.98 & 0.98 & 1.31 & 1.24 & 1.24 & 1.18 & 1.49 & 1.28 & 1.26 & 1.18 \\
    Halfsort       & 1.10 & 0.99 & 1.01 & 1.02 & 1.03 & 0.97 & 0.98 & 0.98 & 1.02 & 1.00 & 0.99 & 0.98 \\
    SmallVec       & 0.99 & 1.02 & 1.03 & 1.05 & 1.01 & 0.98 & 0.99 & 0.98 & 1.02 & 0.99 & 0.99 & 0.99 \\
    $k{=}1$ fast   & 1.04 & 0.91 & 0.91 & 0.94 & 1.03 & 0.96 & 0.98 & 0.98 & 1.01 & 0.99 & 0.99 & 0.99 \\
    Triangle       & 1.01 & 1.01 & 1.01 & 1.03 & 0.98 & 0.98 & 0.98 & 0.99 & 1.03 & 0.99 & 0.99 & 0.99 \\
    All off        & 1.06 & 0.93 & 0.92 & 0.91 & 1.33 & 1.19 & 1.20 & 1.15 & 1.58 & 1.30 & 1.37 & 1.44 \\
    \bottomrule
  \end{tabular}
\end{table}

Early-exit has no effect on corel at any~$k$ ($0.98$--$0.99\times$). On
covtype and mnist its benefit shrinks as $k$ grows, from $1.31\times$ and
$1.49\times$ at $k{=}1$ to $1.18\times$ on both at $k{=}50$, consistent with a
larger $k$ widening the pruning radius, so that fewer candidate distances
exceed the bound and can be cut short.
The other optimizations (halfsort, SmallVec, triangle, $k{=}1$ fast path) are
within a few percent of $1.0\times$ on covtype and mnist, confirming that
traversal-level micro-optimizations have diminishing returns when
distance computation dominates. The one sizable entry, corel's $k{=}1$ fast
path at $0.91$--$0.94\times$ for $k{\ge}5$, comes from a path that is never taken
at $k{>}1$ and so is a build-to-build layout effect, not an effect of the
optimization (Section~\ref{sec:ablation}).

%% ===========================================================================
\FloatBarrier
\clearpage
\section{Distance Computation Tables}
\label{sec:app_dist_counts}
%% ===========================================================================

Table~\ref{tab:app_dist_full} presents full distance computation counts across
all datasets and variants for the all-NN benchmark at $k{=}1$.

\begin{table}[h]
  \caption{Distance computations for all-NN ($k{=}1$, base~1.3, dual-tree mode for
    Packed/NA\,Pk/NA, single-tree all-NN for Simplified/Parallel, the reference
    implementation's batch query for C++) across all
    12~datasets. Values in billions (B), millions (M) or thousands (K). The ablation in
    Section~\ref{sec:ablation} measures the same dual-tree Packed mode, so its
    baseline counts are the Packed column here.}
  \label{tab:app_dist_full}
  \centering
  \small
  \setlength{\tabcolsep}{3pt}
  \begin{tabular}{l rrrrrr}
    \toprule
    Dataset & Packed & NA Pk & NA & Simpl. & Parallel & C++ \\
    \midrule
    iris       & 5.27K & 5.22K & 5.22K & 6.62K & 7.63K & 2.63K \\
    wine       & 3.75K & 4.25K & 4.25K & 9.47K & 11.7K & 2.64K \\
    glass      & 12.9K & 11.9K & 11.9K & 19.5K & 22.2K & 5.63K \\
    bupa       & 36.7K & 37.6K & 37.6K & 44.6K & 60.9K & 17.9K \\
    pima       & 77.1K & 88.8K & 88.8K & 208K & 192K & 41.9K \\
    ionosphere & 43.2K & 50K & 50K & 45.8K & 59.7K & 18.1K \\
    optdigits  & 13.2M & 13.2M & 13.2M & 2.31M & 5.57M & 5.23M \\
    pendigits  & 5.08M & 5.85M & 5.85M & 5.07M & 7.16M & 2.81M \\
    letter     & 42.7M & 50.1M & 50.1M & 22.3M & 26.9M & 18.6M \\
    corel      & 160M & 197M & 197M & 50.5M & 104M & 76.6M \\
    covtype    & 230M & 483M & 483M & 1.55B & 1.7B & 136M \\
    mnist      & 1.87B & 1.96B & 1.96B & 312M & 510M & 697M \\
    \bottomrule
  \end{tabular}
\end{table}

%% ===========================================================================
\FloatBarrier
\clearpage
\section{Batch Single-Tree and Dual-tree Algorithm Diagrams}
\label{sec:app_batch_diagram}
%% ===========================================================================

\begin{figure}[h]
  \centering
  \begin{tikzpicture}[
    qnode/.style={circle, draw=blue!70, fill=blue!10, minimum size=6mm,
                  inner sep=0pt, font=\footnotesize\bfseries},
    rnode/.style={rectangle, draw=red!70, fill=red!10, minimum size=5mm,
                  inner sep=1pt, font=\footnotesize},
    pruned/.style={rectangle, draw=red!30, fill=red!3, minimum size=5mm,
                  inner sep=1pt, font=\footnotesize, text=red!40},
    coverset/.style={draw=gray, rounded corners=2pt, dashed, inner sep=3pt},
    scalelabel/.style={font=\footnotesize\itshape, text=gray},
    arr/.style={-{Stealth[length=2mm]}, thick},
    inheritarr/.style={-{Stealth[length=2mm]}, thick, blue!60},
    prunearr/.style={-{Stealth[length=2mm]}, thick, red!50, dashed},
    every node/.append style={font=\small},
  ]
    % Scale labels (far left, clear of brace)
    \node[font=\footnotesize\itshape, text=gray, anchor=east] at (-4.5, 0) {scale $s$};
    \node[font=\footnotesize\itshape, text=gray, anchor=east] at (-4.5, -2.0) {scale $s{-}1$};
    \node[font=\footnotesize\itshape, text=gray, anchor=east] at (-4.5, -4.0) {scale $s{-}2$};

    % Horizontal dotted lines for scale levels
    \draw[gray, densely dotted] (-3.0, -1.0) -- (5.6, -1.0);
    \draw[gray, densely dotted] (-3.0, -3.0) -- (5.6, -3.0);

    % Brace: amortization annotation
    \draw[decorate, decoration={brace, amplitude=4pt, mirror}, thick, gray]
      (-2.2, -4.2) -- (-2.2, -0.2)
      node[midway, anchor=east, xshift=-5pt,
           font=\scriptsize, text=gray, text width=1.5cm, align=center]
      {pruning at~$q$ saves work for all of~$q$'s subtree};

    % Query tree (left)
    \node[qnode] (q) at (0, 0) {$q$};
    \node[qnode] (qc1) at (-1.0, -2.0) {$q_1$};
    \node[qnode] (qc2) at (1.0, -2.0) {$q_2$};
    \node[qnode] (qc1a) at (-1.5, -4.0) {$q_{1a}$};
    \node[qnode] (qc1b) at (-0.5, -4.0) {$q_{1b}$};

    \draw[arr, blue!60] (q) -- (qc1);
    \draw[arr, blue!60] (q) -- (qc2);
    \draw[arr, blue!60] (qc1) -- (qc1a);
    \draw[arr, blue!60] (qc1) -- (qc1b);

    % Cover sets (right) at scale s: full set
    \node[rnode] (r1) at (3.3, 0) {$r_1$};
    \node[rnode] (r2) at (4.1, 0) {$r_2$};
    \node[rnode] (r3) at (4.9, 0) {$r_3$};
    \node[coverset, fit=(r1)(r2)(r3), label={[font=\scriptsize]above:cover set}] (cs0) {};

    % Cover sets at scale s-1: filtered for q_1
    \node[rnode] (r1b) at (3.3, -2.0) {$r_1$};
    \node[rnode] (r2b) at (4.1, -2.0) {$r_2$};
    \node[coverset, fit=(r1b)(r2b),
      label={[font=\scriptsize]above right:filtered for $q_1$}] (cs1) {};

    % r3 pruned
    \node[pruned] (r3x) at (5.3, -2.0) {$r_3$};

    % Arrows
    \draw[inheritarr] (cs0.south west) -- (cs1.north west)
      node[midway, right, font=\scriptsize, text=blue!70] {copy + filter};
    \draw[prunearr] (r3.south) -- (r3x.north)
      node[midway, right, font=\scriptsize, text=red!60] {pruned};

  \end{tikzpicture}
  \caption{Batch single-tree query.  The query tree (blue, left) is traversed
    top-down by scale level.  At each query split, children inherit
    \emph{filtered copies} of the parent's reference cover sets (red, right).
    A reference node pruned at~$q$ is never seen by any descendant, amortizing
    the pruning decision across all queries in~$q$'s subtree.}
  \label{fig:batch_st}
\end{figure}
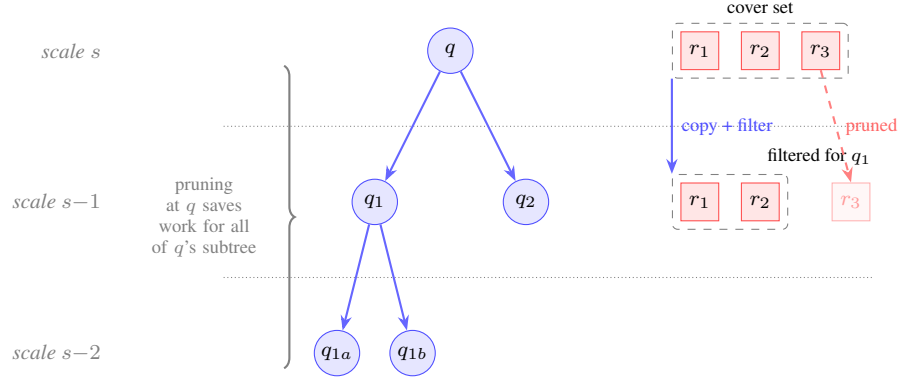

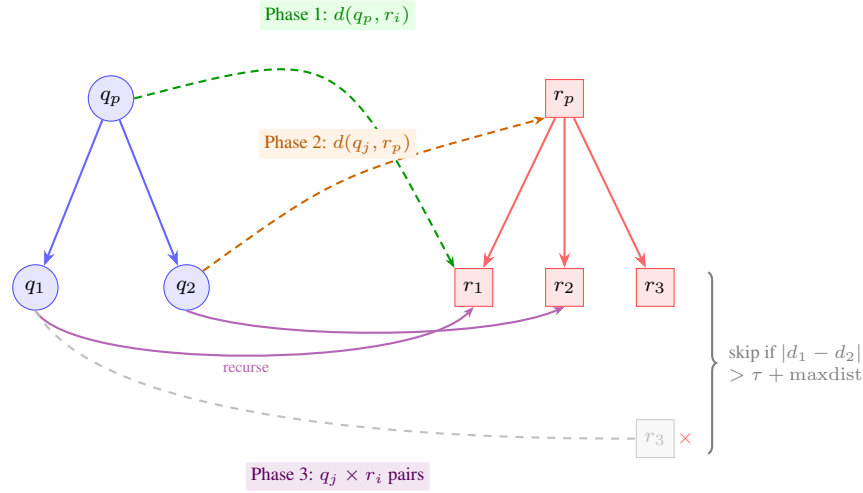
\begin{figure}[h]
  \centering
  \begin{tikzpicture}[
    qnode/.style={circle, draw=blue!70, fill=blue!10, minimum size=6mm,
                  inner sep=0pt, font=\footnotesize\bfseries},
    rnode/.style={rectangle, draw=red!70, fill=red!10, minimum size=5mm,
                  inner sep=1pt, font=\footnotesize},
    pruned/.style={rectangle, draw=gray!40, fill=gray!5, minimum size=5mm,
                  inner sep=1pt, font=\footnotesize, text=gray!50},
    phaselabel/.style={font=\scriptsize, rounded corners=1pt, inner sep=2pt},
    arr/.style={-{Stealth[length=2mm]}, thick},
    every node/.append style={font=\small},
  ]
    % Query tree (left side)
    \node[qnode] (qp) at (0, 0) {$q_p$};
    \node[qnode] (qc1) at (-1.0, -2.5) {$q_1$};
    \node[qnode] (qc2) at (1.0, -2.5) {$q_2$};
    \draw[arr, blue!60] (qp) -- (qc1);
    \draw[arr, blue!60] (qp) -- (qc2);

    % Reference tree (right side)
    \node[rnode] (rp) at (6.0, 0) {$r_p$};
    \node[rnode] (rc1) at (4.8, -2.5) {$r_1$};
    \node[rnode] (rc2) at (6.0, -2.5) {$r_2$};
    \node[rnode] (rc3) at (7.2, -2.5) {$r_3$};
    \draw[arr, red!60] (rp) -- (rc1);
    \draw[arr, red!60] (rp) -- (rc2);
    \draw[arr, red!60] (rp) -- (rc3);

    % Phase 1: q_p vs r_children (green) — arcs high above
    \draw[-{Stealth[length=1.5mm]}, thick, green!60!black, densely dashed]
      (qp.east) .. controls (3.0, 0.7) .. (rc1.north west);
    \node[phaselabel, fill=green!10, text=green!50!black]
      at (3.0, 1.1) {Phase 1: $d(q_p, r_i)$};

    % Phase 2: q_children vs r_p (orange) — arcs through mid region
    \draw[-{Stealth[length=1.5mm]}, thick, orange!80!black, densely dashed]
      (qc2.north east) .. controls (3.0, -1.0) .. (rp.south west);
    \node[phaselabel, fill=orange!10, text=orange!70!black]
      at (3.0, -0.6) {Phase 2: $d(q_j, r_p)$};

    % Phase 3: q_child x r_child pairs (violet)
    \node[phaselabel, fill=violet!10, text=violet!70!black] (p3label)
      at (3.0, -5.0) {Phase 3: $q_j \times r_i$ pairs};

    % Phase 3 arrows — curved below the child row to avoid crossing q2
    % q1 → r1 (recurse): arcs down to y=-3.5
    \draw[-{Stealth[length=1.5mm]}, thick, violet!60]
      (qc1.south) .. controls (-0.5, -3.6) and (4.0, -3.6) .. (rc1.south)
      node[pos=0.5, below, font=\tiny, text=violet!60] {recurse};
    % q2 → r2 (recurse): arcs down to y=-3.2 (shallower, distinct path)
    \draw[-{Stealth[length=1.5mm]}, thick, violet!60]
      (qc2.south) .. controls (2.0, -3.2) and (5.0, -3.2) .. (rc2.south);

    % Pruned pair: q1 → r3 (gray dashed, arcs below)
    \node[pruned] (rc3g) at (7.2, -4.5) {$r_3$};
    \draw[thick, gray!50, dashed]
      (qc1.south) .. controls (0.0, -4.5) and (5.5, -4.5) .. (rc3g.west);
    \node[font=\scriptsize, text=red!60] at (7.6, -4.5) {$\times$};

    % Triangle inequality annotation
    \draw[decorate, decoration={brace, amplitude=3pt}, thick, gray]
      (7.9, -2.3) -- (7.9, -4.7)
      node[midway, right=4pt, font=\scriptsize, text=gray, align=left]
      {skip if $|d_1 - d_2|$\\$> \tau + \mathrm{maxdist}$};

  \end{tikzpicture}
  \caption{Dual-tree traversal.  At each internal pair $(q_p, r_p)$, three
    phases execute: Phase~1 computes $d(q_p, r_i)$ for each reference child;
    Phase~2 computes $d(q_j, r_p)$ for each query child; Phase~3 evaluates all
    $q_j \times r_i$ pairs, using triangle inequality bounds from Phases~1--2
    to prune without additional distance computations.  Pruned pairs (gray) skip
    entire subtrees.  The quadratic $|\text{children}_q| \times |\text{children}_r|$
    expansion in Phase~3 is the main per-pair cost of the traversal.}
  \label{fig:dual_tree}
\end{figure}

%% ===========================================================================
\FloatBarrier
\clearpage
\section{All-NN \texorpdfstring{$k{=}10$}{k=10} Results}
\label{sec:app_allnn_k10}
%% ===========================================================================

\begin{table}[!h]
  \caption{All-nearest-neighbor query time (ms, $k{=}10$, base~1.3).
    Companion to Table~\ref{tab:allnn} ($k{=}1$).
    Median [95\% CI] over 20 independent SLURM allocations.
    Best result per row in \textbf{bold}.
    Values ${\ge}10^3$ are abbreviated K/M.}
  \label{tab:allnn_k10}
  \centering
  \scriptsize
  \setlength{\tabcolsep}{1.5pt}
  \begin{tabular}{l@{\hspace{3pt}}c rr r rrr}
    \toprule
    & & \multicolumn{3}{c}{\rustknn{}} & C++ & mlpack & mlpack \\
    \cmidrule(lr){3-5}
    Dataset & $d$ & Dual & Single & Batch & Ref & Single & Dual \\
    \midrule
    iris & 4 & 1.33 [1.15, 1.37] & 0.94 [0.77, 0.97] & 1.46 [1.21, 1.48] & \textbf{0.93 [0.80, 0.94]} & 2.67 [2.30, 2.68] & 2.69 [2.38, 2.72] \\
    wine & 13 & 1.34 [1.11, 1.39] & 1.44 [1.18, 1.51] & 1.69 [1.40, 1.70] & \textbf{1.10 [0.91, 1.16]} & 3.27 [2.67, 3.50] & 2.33 [2.01, 2.53] \\
    glass & 9 & 2.83 [2.50, 3.28] & 2.46 [2.14, 2.79] & 2.86 [2.50, 3.28] & \textbf{1.70 [1.51, 1.91]} & 5.51 [4.57, 5.94] & 5.05 [4.17, 5.31] \\
    bupa & 6 & 6.46 [5.66, 6.88] & 5.64 [4.80, 5.85] & 5.44 [4.80, 5.82] & \textbf{3.65 [3.42, 4.09]} & 11.4 [10.7, 13.6] & 10.3 [9.51, 12.6] \\
    pima & 8 & 12.3 [10.5, 13.7] & 20.0 [17.2, 22.5] & 12.3 [10.9, 13.4] & \textbf{7.50 [6.81, 7.55]} & 30.4 [28.7, 31.2] & 23.6 [23.4, 25.1] \\
    ionosphere & 34 & 6.49 [6.34, 6.88] & 7.97 [7.93, 8.47] & 7.39 [7.15, 7.80] & \textbf{4.77 [4.52, 5.08]} & 11.5 [10.5, 12.0] & 10.8 [9.74, 11.2] \\
    optdigits & 64 & 935 [837, 1{,}184] & 772 [736, 902] & 987 [969, 1{,}143] & \textbf{728 [716, 785]} & 2.13K [2.1, 2.24] & 2.33K [2.29, 2.44] \\
    pendigits & 16 & 300 [299, 330] & 383 [382, 409] & 372 [371, 396] & \textbf{260 [256, 284]} & 1.28K [1.26, 1.36] & 949 [945, 1{,}012] \\
    letter & 16 & 1.9K [1.89, 1.99] & 1.9K [1.89, 2.01] & 3.05K [3.03, 3.24] & \textbf{1.71K [1.71, 1.79]} & 9.19K [8.89, 9.54] & 9.95K [9.8, 10.4] \\
    corel & 32 & \textbf{7.58K [7.27, 7.86]} & 9.88K [8.22, 10.2] & 23.6K [19, 24] & 14.5K [12.7, 14.9] & 49.6K [41.3, 50.3] & 73.2K [56.1, 75.2] \\
    covtype & 54 & \textbf{25.2K [22.1, 25.7]} & 185K [160, 190] & 121K [103, 123] & 55.1K [49.8, 55.8] & 232K [214, 232] & 177K [156, 180] \\
    mnist & 784 & \textbf{886K [766, 950]} & 2.90M [2.62, 3.96] & 2.98M [2.70, 3.76] & 3.20M [3.14, 3.51] & 5.29M [5.21, 5.34] & 4.25M [3.20, 5.05] \\
    \bottomrule
  \end{tabular}
\end{table}

%% ===========================================================================
\FloatBarrier
\clearpage
\section{KD-Tree Baseline: Ablation and Cross-Codebase Comparison}
\label{sec:app_kdtree}
%% ===========================================================================

This appendix supports Section~\ref{sec:beyond_ct}. The KD-tree is a
median-split tree implemented inside \rustknn{} (\texttt{src/kdtree/}),
reusing the \texttt{Distance} trait (including \texttt{distance\_with\_bound},
the early-exit kernel), the same $k$-NN candidate structure, the same
thread-local distance counter, and the same loader and timing code as the cover
trees. Correctness was verified against brute-force search on every dataset
(maximum coordinate discrepancy $10^{-16}$). Because it shares the kernel, it
runs the same five feature-flag ablations.

\begin{table}[h]
  \caption{Cross-structure and cross-codebase comparison (all-NN, $k{=}1$, ms).
    ``KD-tree'' and the two cover-tree columns share one codebase, kernel, and
    candidate structure, so their differences are attributable to the index
    itself. \texttt{kiddo} is an independent Rust KD-tree crate
    (median [95\% CI] over 20 allocations) and isolates what changes when only
    the codebase differs. KD-tree and cover-tree columns come from the same
    dedicated single-allocation run, so that comparison is controlled same-run; the cover-tree
    values agree with Table~\ref{tab:allnn} within 2\% on covtype and within
    7--19\% on the other datasets, the differences reflecting
    cross-campaign variance (Section~\ref{sec:experiments}).}
  \label{tab:kdtree}
  \centering
  \small
  \setlength{\tabcolsep}{4pt}
  \begin{tabular}{l c rrr r}
    \toprule
    & & \multicolumn{3}{c}{Same codebase} & Independent \\
    \cmidrule(lr){3-5} \cmidrule(lr){6-6}
    Dataset & $d$ & KD-tree & CT single & CT dual & \texttt{kiddo} \\
    \midrule
    pendigits & 16  & 71.9 & 171 & 199 & 7.77 [6.27, 9.80] \\
    corel     & 32  & 254 & 3{,}248 & 5{,}702 & 59.1 [55.3, 70.8] \\
    covtype   & 54  & 4{,}465 & 100{,}590 & 13{,}722 & 12{,}846 [12{,}777, 12{,}943] \\
    mnist     & 784 & 4.09M & 334{,}487 & 628{,}683 & 8.31M [8.29M, 8.34M] \\
    \bottomrule
  \end{tabular}
\end{table}

\begin{table}[h]
  \caption{KD-tree optimization ablation: slowdown ($\times$ baseline) when
    each flag is disabled ($k{=}1$, all-NN). Directly comparable to the
    cover-tree ablation in Table~\ref{tab:ablation}, since both share the
    distance kernel and candidate structure. Median of five runs in a single
    allocation.}
  \label{tab:app_kdtree_ablation}
  \centering
  \small
  \begin{tabular}{l ccccc}
    \toprule
    Disabled optimization & pima & pendigits & corel & covtype & mnist \\
                          & ($d{=}8$) & ($d{=}16$) & ($d{=}32$) & ($d{=}54$) & ($d{=}784$) \\
    \midrule
    Early-exit distance & 0.90 & 0.98 & 0.99 & 0.97 & 1.11 \\
    Halfsort            & 0.62 & 1.04 & 1.00 & 1.00 & 1.01 \\
    SmallVec            & 1.02 & 1.01 & 0.99 & 0.98 & 1.00 \\
    $k{=}1$ fast path   & 1.09 & 1.00 & 1.00 & 1.01 & 1.00 \\
    Triangle pre-filter & 0.70 & 1.00 & 1.00 & 0.99 & 1.02 \\
    All off             & 0.98 & 0.98 & 0.96 & 1.02 & 1.11 \\
    \bottomrule
  \end{tabular}
\end{table}

The pattern follows the cover-tree result in the way one would predict from
first principles: early-exit distance is the only flag with an effect beyond
noise, and only at high dimension ($1.11\times$ at $d{=}784$; ${\approx}1\times$
at $d \le 54$), because it acts on per-distance cost, which is what scales
with~$d$. The traversal-level flags designed around cover-tree
invariants---halfsort and the triangle pre-filter---sit at ${\approx}1\times$
on a structure whose traversal they do not describe. As with the cover trees,
the pima column is noise-dominated and we draw nothing from it.

Table~\ref{tab:app_kdtree_dist} gives the KD-tree's distance counts, which make
the curse-of-dimensionality crossover explicit: on mnist the KD-tree issues
594M distance computations (each up to 784-dimensional), versus 312M for the
simplified cover tree in single-tree mode (Table~\ref{tab:app_dist_full}),
because axis-aligned box pruning stops discriminating at high~$d$.

\begin{table}[h]
  \caption{KD-tree all-NN metric evaluations ($k{=}1$), all 12 datasets.
    Cover-tree counts for all 12 datasets appear in
    Appendix~\ref{sec:app_dist_counts}.}
  \label{tab:app_kdtree_dist}
  \centering
  \small
  \begin{tabular}{l r r}
    \toprule
    Dataset & $d$ & KD-tree distance evaluations \\
    \midrule
    iris       & 4   & 1{,}126 \\
    wine       & 13  & 1{,}207 \\
    glass      & 9   & 1{,}534 \\
    bupa       & 6   & 2{,}914 \\
    pima       & 8   & 7{,}605 \\
    ionosphere & 34  & 3{,}773 \\
    optdigits  & 64  & 417{,}599 \\
    pendigits  & 16  & 274{,}892 \\
    letter     & 16  & 2{,}111{,}157 \\
    corel      & 32  & 566{,}391 \\
    covtype    & 54  & 10{,}982{,}600 \\
    mnist      & 784 & 594{,}194{,}887 \\
    \bottomrule
  \end{tabular}
\end{table}

For the cross-codebase column we used \texttt{kiddo}~5.3.2, an actively
maintained independent Rust KD-tree crate, via its \texttt{ImmutableKdTree}
with \texttt{nearest\_n::<SquaredEuclidean>}. It was run under the same
20-allocation protocol as the main tables (all-NN, $k{=}1$);
because \texttt{kiddo} is compiled as part of the same Rust project, it
inherits the same \texttt{RUSTFLAGS} (including
\texttt{target-cpu=native}), so the ISA disparity noted in
Appendix~\ref{sec:exp_impl} does not apply to this comparison.
\texttt{kiddo} necessarily uses its own distance kernel and candidate
structure, which is why it cannot run our ablation flags and why we
implemented an in-harness tree for the controlled comparison: an external
library alone would reintroduce exactly the cross-codebase confounding this
paper is about. Used together, the two answer different questions---the
in-harness tree isolates the index, and \texttt{kiddo} quantifies what the
codebase alone contributes.

%% ===========================================================================
\FloatBarrier
\clearpage
\section{Naive Search with Partial Distance Computation}
\label{sec:app_naive}
%% ===========================================================================

A reasonable question about any index is whether it earns its complexity. We
added a \texttt{naive} variant---a linear scan---to the same harness, using the
same early-exit (``partial distance'') Euclidean kernel and the same distance
counter as every other implementation. Building it with
\texttt{no-early-exit} yields the full-distance twin. Both perform identical
$n^2$ distance \emph{calls}, so any runtime difference is purely the
per-call dimension work that early exit skips---this isolates the
partial-distance effect on the simplest possible baseline.

\begin{table}[h]
  \caption{Naive all-NN ($k{=}1$): full vs.\ partial distance computation.
    Both columns issue the same number of distance calls; the difference is
    per-call work avoided. Median of three runs in one allocation; on covtype, the six runs (three per
    variant) ran concurrently on one node so both variants saw the same load.}
  \label{tab:app_naive}
  \centering
  \small
  \begin{tabular}{l r rr r r}
    \toprule
    Dataset & $d$ & Full (ms) & Partial (ms) & Speedup & $n^2$ distance calls \\
    \midrule
    pima       & 8   & 5.2 & 4.6 & 1.14$\times$ & 589{,}824 \\
    ionosphere & 34  & 4.3 & 3.0 & 1.41$\times$ & 123{,}201 \\
    pendigits  & 16  & 1{,}019 & 1{,}110 & 0.92$\times$ & 120{,}824{,}064 \\
    optdigits  & 64  & 1{,}047 & 1{,}065 & 0.98$\times$ & 31{,}584{,}400 \\
    letter     & 16  & 3{,}519 & 3{,}678 & 0.96$\times$ & 400{,}000{,}000 \\
    corel      & 32  & 51{,}901 & 20{,}603 & 2.52$\times$ & 1{,}424{,}987{,}001 \\
    covtype    & 54  & 34{,}477{,}113 & 21{,}721{,}950 & 1.59$\times$ & 337{,}574{,}944{,}144 \\
    mnist      & 784 & 4{,}206{,}345 & 2{,}551{,}888 & 1.65$\times$ & 3{,}600{,}000{,}000 \\
    \bottomrule
  \end{tabular}
\end{table}

The partial-distance benefit depends on the dataset more than on its
dimension: it is large on corel ($2.52\times$, $d{=}32$), mnist
($1.65\times$, $d{=}784$) and covtype ($1.59\times$, $d{=}54$), but negligible or slightly negative on pendigits,
letter, and optdigits ($0.92$--$0.98\times$, $d{=}16$--$64$), where the
branch checks cost about as much as the work they skip. Corel is the
instructive case: early exit is worth $2.5\times$ in the linear scan but
nothing in the cover tree (Table~\ref{tab:ablation}), consistent with the tree
having already pruned most of the far candidates whose distance computations
early exit would otherwise cut short. The benefit of a kernel-level
optimization thus depends on the structure that calls the kernel, not only on
the data.

The absolute numbers also answer the ``is an index worth it?'' question
directly. Naive all-NN on mnist requires $3.6{\times}10^9$ distance calls
(43--70 minutes) and on covtype $3.4{\times}10^{11}$ (6--10 hours), against
milliseconds to minutes for the trees. At the same time, on the smallest
datasets the naive scan is competitive in absolute terms---another reason the
tiny-dataset results in this paper should not drive implementation decisions.

%% ===========================================================================
\FloatBarrier
\clearpage
\section{Synthetic Manifolds with Known Intrinsic Dimension}
\label{sec:app_synthetic}
%% ===========================================================================

The cover tree's $O(c^6 \log n)$ query bound is stated in terms of the
\emph{doubling} dimension, not the ambient dimension, but on real datasets the
intrinsic dimension is unknown, so the distinction cannot be tested directly.
We therefore generated synthetic data where it is known by construction:
$n{=}20{,}000$ points sampled on a $d$-dimensional latent manifold, then
linearly embedded in ambient dimension $D \ge d$: latent coordinates are uniform
on $[0,1]^d$, the embedding is a fixed random $D \times d$ matrix with
$\mathcal{N}(0,1)$ entries, and isotropic Gaussian noise ($\sigma{=}0.01$) puts
points near rather than exactly on the manifold, with a fixed seed per
configuration. These run through the harness unchanged. Two sweeps separate the two dimensions.

\begin{table}[h]
  \caption{Synthetic manifolds ($n{=}20{,}000$, all-NN $k{=}1$). Cover-tree
    cost tracks the \emph{intrinsic} dimension~$d$ and does not grow with the
    \emph{ambient} dimension~$D$. KD-tree distance counts are shown for
    contrast: they are near-identical (267K $\pm$ 5) across all
    configurations. Median of three runs in a single allocation.}
  \label{tab:app_synthetic}
  \centering
  \small
  \setlength{\tabcolsep}{4pt}
  \begin{tabular}{ll rr r}
    \toprule
    Sweep & Configuration & CT time (ms) & CT distance comps. & KD distance comps. \\
    \midrule
    \multirow{3}{*}{A: vary $d$ ($D{=}100$)}
      & $d{=}2$  & 669 & 5{,}404{,}434 & 267{,}253 \\
      & $d{=}5$  & 1{,}203 & 8{,}782{,}805 & 267{,}248 \\
      & $d{=}10$ & 3{,}182 & \textbf{22{,}633{,}024} & 267{,}248 \\
    \midrule
    \multirow{3}{*}{B: vary $D$ ($d{=}5$)}
      & $D{=}20$  & 501 & 9{,}765{,}110 & 267{,}248 \\
      & $D{=}50$  & 918 & 10{,}985{,}580 & 267{,}248 \\
      & $D{=}100$ & 1{,}203 & \textbf{8{,}782{,}805} & 267{,}248 \\
    \bottomrule
  \end{tabular}
\end{table}

Sweep~A raises the intrinsic dimension at fixed ambient dimension: cover-tree
distance computations rise steeply, $5.4$M${\to}8.8$M${\to}22.6$M, a
$4.2\times$ increase for a $5\times$ increase in~$d$.
Sweep~B raises the ambient dimension $5\times$ at fixed intrinsic dimension:
distance computations do \emph{not} rise monotonically
($9.8$M${\to}11.0$M${\to}8.8$M---variation is about 25\%, consistent with
different random embedding matrices across configurations rather than an
intrinsic-dimension effect). This is
direct empirical support for the claim that cover-tree query cost is governed
by intrinsic rather than ambient dimensionality, which the theory asserts but
which, to our knowledge, has not previously been isolated experimentally.

Wall-clock behaves differently from distance count in Sweep~B, and the
difference is informative: cover-tree time still rises with~$D$
($501{\to}1{,}203$\,ms) even though the number of distance computations does not,
because each computation is $D$-dimensional. This is the
distance-count-versus-runtime decoupling discussed in
Section~\ref{sec:ablation}, and it is why we report both throughout: a paper
reporting only distance counts here would conclude, incorrectly, that ambient
dimension is free.

%% ===========================================================================
\FloatBarrier
\clearpage
\section{Literature Survey: Evaluation Practices in Metric Space Search}
\label{sec:app_litsurvey}
%% ===========================================================================

To contextualize our experimental contributions, we surveyed 61~papers from
the metric space nearest neighbor search literature and classified each
on four evaluation-practice dimensions that our work addresses:
\begin{enumerate}
  \item \textbf{Micro-optimizations} --- Does the paper discuss
    implementation-level micro-optimizations (e.g., SIMD, cache layout,
    memory coalescing, bit tricks) as opposed to purely algorithmic
    improvements?
  \item \textbf{Runtime \& distance counts} --- Does the paper measure
    \emph{both} wall-clock runtime and the number of distance
    computations (or avoided/pruned computations) in its experiments?
  \item \textbf{Batch vs.\ single query} --- Does the paper compare or
    test between bulk/batch querying and single-item-at-a-time querying?
  \item \textbf{Query from set vs.\ held-out} --- Does the paper
    explicitly differentiate or test between queries drawn from the
    indexed dataset versus new held-out points not in the data structure?
\end{enumerate}

Table~\ref{tab:litsurvey} presents the full classification.
A~``1'' indicates the paper reasonably addresses the criterion; ``0''
indicates it does not.

\begin{longtable}{ccccc}
\caption{Literature classification of 61 metric space search papers on
  four evaluation-practice criteria.}\label{tab:litsurvey}\\
\toprule
Paper & \rotatebox{70}{Micro-opt.} & \rotatebox{70}{\shortstack{Runtime \&\\dist.\ counts}} & \rotatebox{70}{\shortstack{Batch vs.\\single}} & \rotatebox{70}{\shortstack{Set vs.\\held-out}} \\
\midrule
\endfirsthead
\caption[]{(continued)}\\
\toprule
Paper & \rotatebox{70}{Micro-opt.} & \rotatebox{70}{\shortstack{Runtime \&\\dist.\ counts}} & \rotatebox{70}{\shortstack{Batch vs.\\single}} & \rotatebox{70}{\shortstack{Set vs.\\held-out}} \\
\midrule
\endhead
\midrule
\multicolumn{5}{r}{\footnotesize\itshape continued on next page}\\
\endfoot
\bottomrule
\endlastfoot
\cite{NIPS2000_1848} & 0 & 0 & 0 & 0 \\
\cite{Li:2015:CWS:2783258.2783406} & 0 & 0 & 0 & 0 \\
\cite{Fukunage1975} & 0 & 0 & 0 & 0 \\
\cite{elkin_new_2023} & 0 & 0 & 0 & 0 \\
\cite{petrov_note_2023} & 0 & 0 & 0 & 0 \\
\cite{moore-tutorial} & 0 & 0 & 0 & 1 \\
\cite{Cayton2012} & 1 & 0 & 1 & 0 \\
\cite{NIPS2004_2666} & 0 & 1 & 0 & 0 \\
\cite{indyk1998approximate} & 0 & 0 & 0 & 0 \\
\cite{Nielsen2009} & 0 & 1 & 0 & 1 \\
\cite{Wu:2017:CWS:3038912.3052598} & 0 & 0 & 0 & 0 \\
\cite{elkin_counterexamples_2022} & 0 & 0 & 0 & 0 \\
\cite{beygelzimer2006cover} & 0 & 0 & 0 & 0 \\
\cite{yianilos1993data} & 0 & 1 & 0 & 0 \\
\cite{Bozkaya:1997:DIH:253260.253345} & 0 & 1 & 0 & 0 \\
\cite{Wu2006} & 1 & 0 & 1 & 1 \\
\cite{Yianilos98} & 0 & 0 & 0 & 0 \\
\cite{curtin_exploiting_2019} & 0 & 0 & 0 & 1 \\
\cite{curtin_fast_2016} & 0 & 0 & 0 & 1 \\
\cite{curtin_fast_2013} & 0 & 0 & 0 & 1 \\
\cite{izbicki2015faster} & 1 & 1 & 0 & 1 \\
\cite{curtin_faster_2015} & 0 & 0 & 1 & 1 \\
\cite{Karger:2002:FNN:509907.510013} & 0 & 0 & 0 & 0 \\
\cite{omohundro1989five} & 0 & 0 & 0 & 0 \\
\cite{Samet:2005:FMM:1076819} & 0 & 0 & 0 & 0 \\
\cite{Kamel:1994:HRI:645920.673001} & 0 & 0 & 0 & 0 \\
\cite{Curtin2015} & 0 & 1 & 1 & 1 \\
\cite{Uhlmann1991a} & 0 & 0 & 0 & 1 \\
\cite{Ioffe:2010:ICS:1933307.1934593} & 0 & 0 & 0 & 0 \\
\cite{Hjaltason2000} & 0 & 0 & 0 & 0 \\
\cite{Bozkaya1999} & 0 & 0 & 0 & 1 \\
\cite{Samet2008} & 0 & 0 & 0 & 0 \\
\cite{NIPS2013_fc8001f8} & 0 & 0 & 0 & 0 \\
\cite{Pauleve2010} & 0 & 0 & 0 & 0 \\
\cite{Andoni2006} & 0 & 0 & 0 & 0 \\
\cite{Yianilos:2000:LLC:338219.338273} & 0 & 0 & 0 & 1 \\
\cite{667} & 0 & 0 & 0 & 1 \\
\cite{Ciaccia:1997:MEA:645923.671005} & 0 & 0 & 0 & 0 \\
\cite{Bentley:1975:MBS:361002.361007} & 0 & 0 & 0 & 0 \\
\cite{Brin1995} & 0 & 0 & 0 & 0 \\
\cite{Andoni2006a} & 0 & 0 & 0 & 0 \\
\cite{clarkson1999nearest} & 0 & 0 & 0 & 0 \\
\cite{Clarkson2002} & 0 & 1 & 0 & 0 \\
\cite{ram_nearest-neighbor_2012}  & 0 & 0 & 0 & 1 \\
\cite{gu_parallel_2022} & 0 & 0 & 1 & 0 \\
\cite{Rajani2015} & 0 & 0 & 0 & 0 \\
\cite{Kim2013} & 1 & 0 & 1 & 0 \\
\cite{Papadopoulos1997}  & 0 & 0 & 0 & 0 \\
\cite{Finkel:1974:QTD:2697709.2697865} & 0 & 0 & 0 & 0 \\
\cite{huang_query-aware_2017} & 0 & 0 & 0 & 1 \\
\cite{Guttman1984} & 0 & 0 & 0 & 0 \\
\cite{Dasgupta:2008:RPT:1374376.1374452} & 0 & 0 & 0 & 0 \\
\cite{Dasgupta:2015:RPT:2756000.2756040} & 0 & 0 & 0 & 0 \\
\cite{Ram:2019:RKN:3292500.3330875} & 1 & 0 & 0 & 1 \\
\cite{Chavez:2001:SMS:502807.502808} & 0 & 0 & 0 & 0 \\
\cite{NIPS2016_6472} & 1 & 0 & 0 & 0 \\
\cite{Moore:2000:AHU:2073946.2073993} & 0 & 1 & 0 & 0 \\
\cite{Samet:1984:QRH:356924.356930} & 0 & 0 & 0 & 0 \\
\cite{curtin2013tree} & 0 & 0 & 0 & 0 \\
\cite{huang_two_2017} & 0 & 0 & 0 & 0 \\
\cite{Kumar:2008:GNN:1479250.1479280} & 1 & 1 & 0 & 0 \\
\end{longtable}

\paragraph{Summary.}
Of the 61~papers surveyed:
\begin{itemize}
  \item Only \textbf{7} (11\%) discuss implementation-level
    micro-optimizations, indicating that the vast majority of metric
    space search work operates at the algorithmic level without examining
    how implementation details interact with algorithmic choices.
  \item Only \textbf{9} (15\%) report \emph{both} wall-clock runtime and
    distance computation counts, despite the fact that these two metrics
    can diverge significantly (e.g., fewer distance computations does not
    guarantee faster runtime when per-computation cost varies or overhead
    dominates).
  \item Only \textbf{6} (10\%) compare batch/bulk querying against
    single-item querying---and these are almost exclusively the dual-tree
    and GPU-parallelism papers.
  \item Only \textbf{16} (26\%) explicitly test with held-out query
    points distinct from the indexed dataset, despite the fact that
    production workloads nearly always involve queries unseen at
    index-construction time.
\end{itemize}
These gaps motivate the four contributions of this paper: our
optimization ablation (addressing gap~1), joint runtime and distance
reporting (gap~2), query-mode comparison (gap~3), and all-NN vs.\
held-out evaluation (gap~4).

\end{document}